\documentclass[10pt, a4paper, logo]{googledeepmind}

\usepackage{times}

\usepackage{hyperref}
\usepackage[leftcaption]{sidecap}

\usepackage{amsmath,amsfonts,bm}
\usepackage{natbib}
\usepackage{fancyhdr}

\def\eqref#1{equation~\ref{#1}}

\def\1{\bm{1}}

\DeclareMathAlphabet{\mathsfit}{\encodingdefault}{\sfdefault}{m}{sl}
\SetMathAlphabet{\mathsfit}{bold}{\encodingdefault}{\sfdefault}{bx}{n}

\usepackage{hyperref}
\usepackage{url}
\AtBeginDocument{\definecolor{tablehl}{HTML}{E8F1FB}}
\usepackage{booktabs}
\usepackage{graphicx}
\usepackage{subcaption}
\usepackage{amssymb}
\usepackage[most]{tcolorbox}
\usepackage[table]{xcolor}
\tcbuselibrary{breakable}
\usepackage{amsmath}
\usepackage{colortbl}

\usepackage{microtype}
\usepackage{color}
\usepackage{wrapfig}
\usepackage{natbib}
\usepackage{array}
\usepackage{textcomp}
\usepackage{stfloats}
\usepackage{float}
\usepackage{verbatim}
\usepackage{algorithm}
\usepackage{algpseudocode}
\usepackage{multirow}
\usepackage{enumitem}

\usepackage{amsfonts}

\usepackage[T1]{fontenc}
\usepackage{etoolbox}
\usepackage[normalem]{ulem}

\usepackage{caption}
\usepackage{makecell}
\usepackage{tabularx}
\let\cite\citep

\usepackage{listings}
\usepackage{pdfpages}
\usepackage{placeins}

\makeatletter
\newcommand{\InlineTableCaption}[2]{
  \refstepcounter{table}\label{#2}
  \@makecaption{\fnum@table}{#1}
}

\newcommand{\AlgStyle}{%
  \small\raggedright
  \algrenewcommand{\algorithmicindent}{1em}%
  \algrenewcommand{\alglinenumber}[1]{\textcolor{black!55}{\scriptsize ##1:}}%
  \algrenewcommand{\algorithmiccomment}[1]{%
    \hspace{0.5em}{\color{black!60}\itshape\(\triangleright\)~##1}}%
}
\newcommand{\AlgNote}[1]{%
  \Statex\hspace{\ALG@tlm}{\color{black!60}\itshape\(\triangleright\)~#1}}
\newcommand{\AlgCont}{\Statex\hspace{\ALG@tlm}\hspace{\algorithmicindent}}
\makeatother

\title{AVA-Encoder: Learning to Represent Videos via Reconstruction-Driven Agentic Self-Evolution}

\author[1,2]{Chuyue Li}
\author[1]{Jinpeng Yu\textsuperscript{\textdagger}}
\author[1,3]{Haozhe Wang}
\author[1,4]{Tian Xueyun}
\author[1,5]{Zhijing Zhang}
\author[1]{Bingnan Li}
\author[2]{Yining Zhao}
\author[6]{Shuqi Gu}
\author[6]{Kan Ren\textsuperscript{*}}
\author[1]{Jiaming Liu\textsuperscript{*}}
\author[1]{Ruihua Huang}
\affil[1]{Qwen Business Unit of Alibaba}
\affil[2]{University of Illinois Urbana-Champaign}
\affil[3]{The Hong Kong University of Science and Technology}
\affil[4]{Institute of Computing Technology}
\affil[5]{Southeast University}
\affil[6]{ShanghaiTech University}
\footnotetext{\textsuperscript{\textdagger}Project Lead. \textsuperscript{*}Co-corresponding authors.\newline Emails: lichuyue0312@gmail.com, jmliu1217@gmail.com.\newline Project Page: \url{https://ava-encoder.github.io/}.}

\begin{abstract}
    Agentic video creation systems remain far from cinematic-grade quality: their underlying models struggle to plan how story, characters, visuals, and sound fit together, and high-quality creation trajectories to learn such planning abilities are scarce. Finished human films embody this knowledge, but in a tightly coupled audiovisual form that agents cannot directly read or manipulate. 
    
    We propose the \textbf{Agentic Video "Auto-Encoder" (AVA-Encoder)}, which works backward from human films, turning them into \textbf{agent-native} video representations with reconstruction itself as the sole supervision. It encodes a film into a Film Knowledge Graph (Film KG) of structured-text nodes, typed dependency edges, and linked multimodal assets, then regenerates the film from this graph with a fixed decoder. Crucially, by distilling reconstruction residuals into textual gradients, AVA-Encoder drives a dual-loop self-evolution: an outer loop autonomously discovers a generalizable encoding strategy entirely from unlabeled videos, while an inner loop refines each video's Film KG at test time.

AVA-Encoder improves reconstruction fidelity by \textbf{20.9} percentage points over the strongest baseline, and its autonomously learned strategy outperforms a human-tuned strategy using \textbf{70\%} fewer prompt tokens.

    Our representations drive three downstream breakthroughs: (1)Film KGs enable \textbf{coordinated graph-based editing} and (2)improve \textbf{one-shot video generation} across four open-source frameworks, while (3)\textbf{fine-tuning agents} on our reverse-engineered trajectories boosts video creation quality by 38.7\%. We release the framework and a dataset of 500 Film KGs with over 27,000 creation trajectories.
    \end{abstract}

\begin{document}
\maketitle

\section{Introduction}
\label{sec:intro}

\begin{figure}[t]
    \centering
    \includegraphics[width=\textwidth]{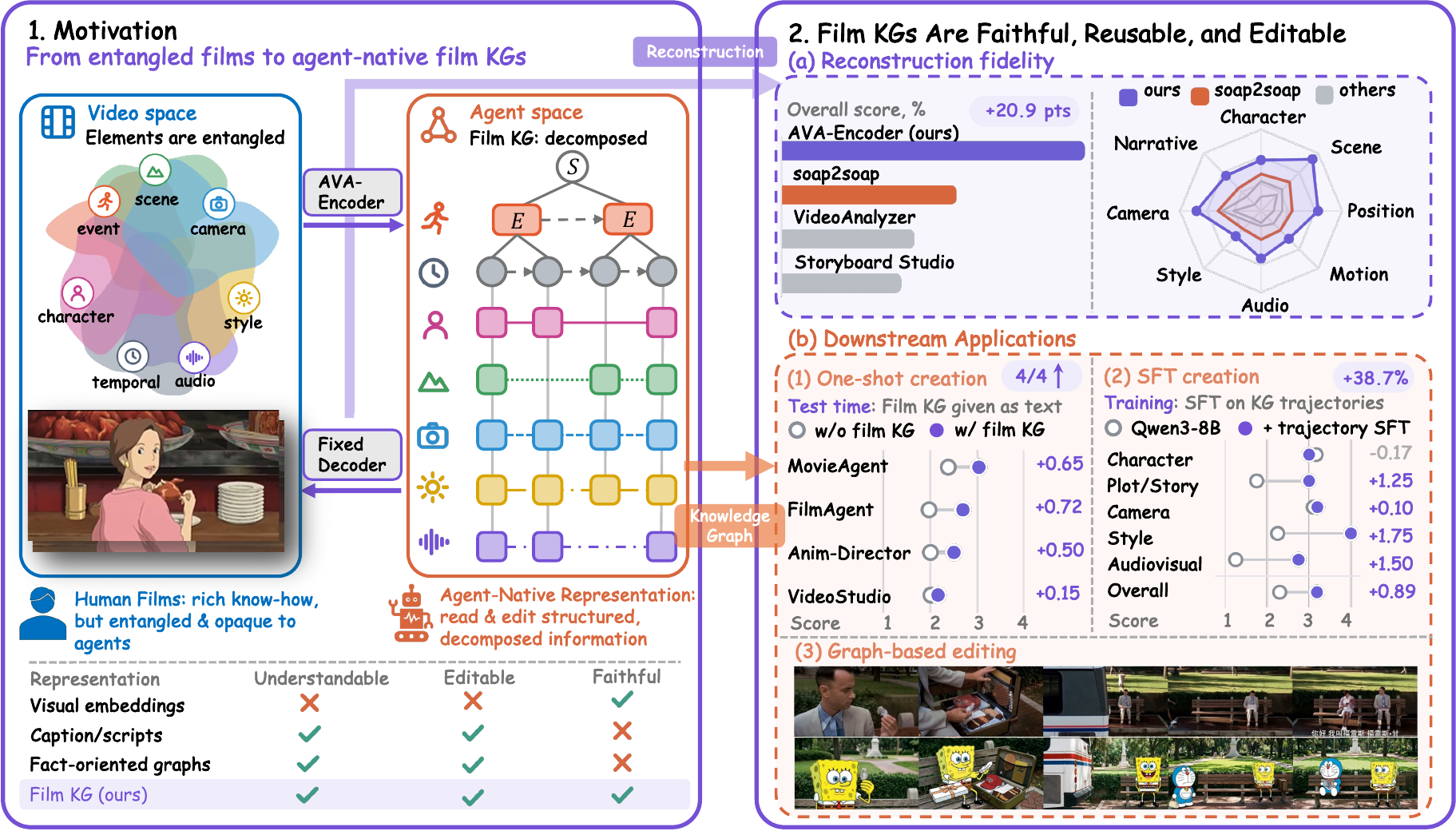}
    \caption{\textbf{Overview of the AVA-Encoder paradigm.} \textbf{1. Motivation:} We map entangled video elements into structured Film KGs, providing an agent-native representation that is understandable, editable, and faithful. \textbf{2.(a) Reconstruction Fidelity:} Film KGs accurately preserve cinematic details, outperforming the strongest baseline by 20.9 percentage points. \textbf{2.(b) Downstream Applications:} The KGs and their reverse-engineered trajectories drive three breakthroughs: one-shot reference creation, trajectory-based agent fine-tuning (+38.7\% quality boost), and coordinated graph-based editing.}
    \label{fig:intro}
\end{figure}

Advances in foundation models~\cite{sora,veo,kling,moviegen,rethinkingcfg,overlaybench} and agentic video creation systems~\cite{filmagent,movieagent,animdirector,videodirectorgpt,visioncreator} have enabled agents to write stories, design keyframes, and generate videos. Yet reliable cinematic-grade creation remains difficult, mainly because their underlying models lack the planning ability to coordinate scripts, characters, shots, and audiovisual elements. Developing this ability is difficult because high-quality trajectories of complete agentic video creation processes are scarce. Meanwhile, cinematic-grade human-made films offer a rich alternative source of filmmaking knowledge~\cite{macro2mind}. However, unlike human filmmakers who learn by directly studying finished films, agents cannot directly use them as explicit creation records.

This limitation reflects a mismatch between video space and agent space: films tightly couple narrative, visual, temporal, and audio information without exposing their underlying relationships~\cite{movienet}, whereas agents use structured text, code, plans, and graphs for retrieval, reasoning, planning, and editing~\cite{wang2025code,ctmai,searchgen}. Bridging this gap requires agent-native representations that make film content and creation dependencies explicit and operable~\cite{agentfirstdata}.\textbf{(Figure~\ref{fig:intro}, 1. Motivation)}

An ideal \textbf{agent-native video representation} should satisfy three requirements: (1) \textbf{understandable to agents}, (2) \textbf{stably-editable for agents}, and (3) \textbf{faithful} enough to preserve the cinematic information for subsequent generation. Existing approaches do not fully meet these requirements: visual embeddings resist direct interpretation and manipulation; textual descriptions often omit cross-modal dependencies; and fact-oriented structured representation prioritize task-relevant facts over the cinematic detail needed for generation (Sec.~\ref{sec:related-agentic_representation}).

\textbf{Research Questions.}This gap poses two fundamental research questions: \emph{(1)How should we design a an \textbf{agent-native video representation} that satisfies these three requirements? (2)And without manual supervision or ground-truth graphs, can an agent \textbf{autonomously learn} to build this representation from raw videos?} 

Solving this requires overcoming three major challenges:
\begin{enumerate}[nosep,leftmargin=1.5em,label=(\arabic*)]
    \item \textbf{Representation Design:} Translating dense multimodal content into structured text risks losing cinematic detail~\cite{verbalts} and failing to preserve complex cross-modal and cross-entity dependencies.
    \item \textbf{Autonomous Optimization Signal:} In the absence of ground-truth representations, formulating a verifiable, trustworthy learning signal for the agent to self-evolve is an open problem.
    \item \textbf{Test-Time Cost and Cross-Video Transfer:} Iterative reconstruction and revision are costly when repeated for every new video. Moreover, an edit to one Film KG improves only that instance, while a shared encoding rule inferred from one video may fail on others. How can the agent reuse what it learns across videos while retaining the ability to correct details specific to a new film?
\end{enumerate}

\begin{figure}[!htbp]
    \centering
    \includegraphics[width=\linewidth]{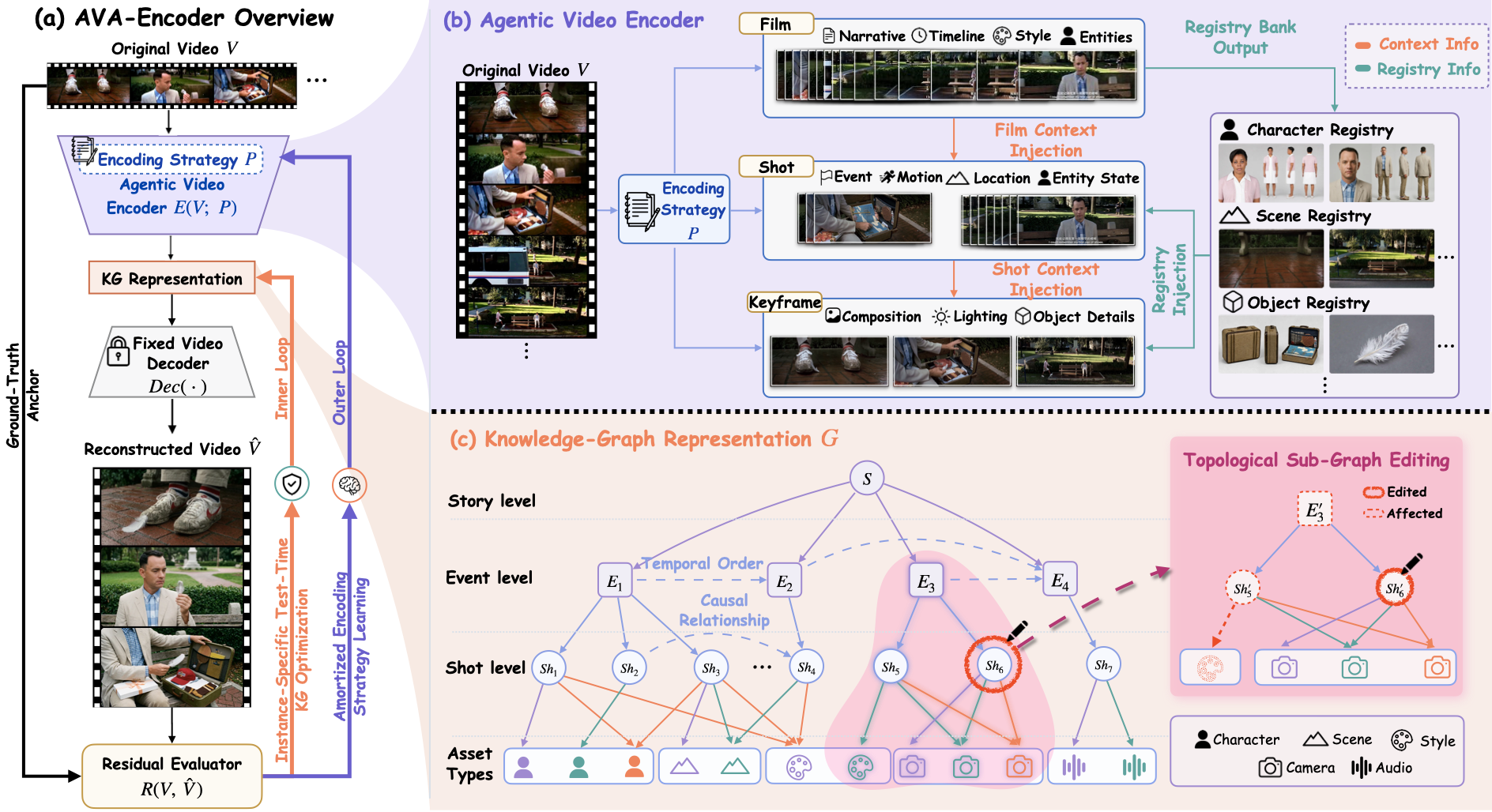}
    \caption{\textbf{Architecture of AVA-Encoder.} (a) Reconstruction-driven, gated \textbf{dual-loop self-evolution}. (b) Hierarchical \textbf{agentic encoder} with inter-level context injection. (c) \textbf{Film KG} nodes, edges, and linked multimodal assets for constrained generation and coordinated editing.}
    \label{fig:pipeline}
\end{figure}

To address these challenges, we propose the \textbf{Agentic Video Auto-Encoder (AVA-Encoder)}, introducing three complementary designs accordingly:
\begin{enumerate}[nosep,leftmargin=1.5em,label=(\arabic*)]
    \item \textbf{Agent-Native Film KG Representation (Fig.~\ref{fig:pipeline}b,c):} We propose the Film Knowledge Graph (KG) to represent video as structured textual nodes and explicit typed edges, built by a hierarchical agentic video encoder. This decouples entangled video elements into agent-readable, editable formats while preserving crucial dependencies.
    \item \textbf{Agentic Auto-Encoding Paradigm (Fig.~\ref{fig:pipeline}a):} We frame agentic self-evolution as an auto-encoding problem. The agent encodes a film into a Film KG, and a fixed decoder reconstructs it. The consistency between the reconstructed and source video serves as a source-derived learning signal that requires no annotated Film KGs~\cite{mae,nerf}, eliminating the need for human annotations or ground-truth graphs.
    \item \textbf{Dual-Loop Cognitive Architecture (Fig.~\ref{fig:dual_loop_intuition}):} Inspired by fast and slow thinking (System 1/System 2), we use gated textual-gradient optimization to first \emph{amortize} expensive search across diverse videos into a fast, generalizable encoding strategy (System 1 prior), reserving instance-specific test-time refinement (System 2 search) only for complex long-tail details.
\end{enumerate}

\textbf{Results.} Extensive experiments validate the AVA-Encoder paradigm:

\begin{itemize}[nosep,leftmargin=1.5em]
    \item \textbf{Video reconstruction as direct evaluation.} Our framework achieves 49.2\% Overall fidelity, outperforming the strongest baseline by 20.9 percentage points \textbf{(Figure~\ref{fig:intro}, 2.a)}. Ablations confirm that our hierarchical Film KG architecture alone yields 42.6\% fidelity even without self-evolution, while our autonomously learned System 1 strategy surpasses a human-tuned strategy using 70\% fewer prompt tokens.

\item \textbf{Dataset and downstream applications.} Beyond reconstruction, we release a dataset of 500 high-quality Film KGs and over 27,000 reverse-engineered creation trajectories. These agent-native representations drive three downstream breakthroughs \textbf{(Figure~\ref{fig:intro}, 2.b)}: the KGs enable (1) coordinated \textbf{graph-based editing} and (2) boost \textbf{one-shot agentic video generation} quality across four open-source systems. (3) Furthermore, \textbf{fine-tuning creative agents} on a subset of 7,863 trajectories improves downstream video creation quality by 38.7\%, mitigating the scarcity of high-quality agentic planning data in video creation.

\end{itemize}

\section{Related Work}
\label{sec:related}

\subsection{Agentic Video Creation Systems}
\label{sec:related-agentic_video_creation}

LLM- and VLM-based systems use planning and tool orchestration for video generation~\cite{videodirectorgpt,vlogger,dreamfactory,mora,visioncreator,animdirector,videostudio,movieagent,filmagent}, editing, and remixing~\cite{soap2soap,comfyui,whatnext}. However, their foundation models struggle to plan complex multimodal cinematic tasks, limiting creation-trajectory quality; the scarcity of high-quality trajectories in this field, in turn, hinders further training to improve planning abilities~\cite{skyrlagent}.

AVA-Encoder instead works backward from human films, converting them into readable, learnable, and editable Film KGs with explicit dependencies, and self-improving the system through reconstruction. Decoding produces aligned intermediate records and rendered outputs, serving as reverse-engineered cinematic-grade creation trajectories for future training~\cite{reverse}, providing a route to learning from abundant high-quality films beyond trajectories current agentic systems can produce.

\subsection{Agent-Native Video Representations}
\label{sec:related-agentic_representation}

Existing video representations include low-level visual forms, such as pixels, latent tokens, and embeddings~\cite{videomae,magvit,videollama}; textual descriptions, such as captions and screenplays~\cite{movienet,longstoryshort,screenwriter}; and structured representations, such as hierarchical memories and graphs~\cite{moviechat,malmm,goldfish,graphvideoagent,egagent}.

Jointly supporting agent readability, reliable editing, and faithful generation remains challenging. (1) \textbf{Low-level forms} retain visual information but are difficult for agents to read, query, or modify directly; (2) \textbf{Textual descriptions} are accessible but often leave entity relationships and cross-modal dependencies implicit. (3) \textbf{Structured representations} support factual retrieval and reasoning, yet need not preserve the cinematic details and generation dependencies required for reconstruction. For example, recording two characters meet in shot eight does not specify their appearance, spatial arrangement, or audiovisual context required for generation.

Success in retrieval and QA~\cite{longva,llavavideo,streamingvu} therefore does not establish generative fidelity. Reconstruction tests whether retained information supports faithful regeneration; AVA-Encoder uses this source-referenced feedback to assess and improve representation fidelity.

\section{Task Definition}
\label{sec:task}

\subsection{Problem Formulation}
\label{sec:task-formulation}

We formulate agent-native video representation learning as \textbf{agentic auto-encoding}, where an agent autonomously learns to encode a video into a structured agentic representation from which a fixed decoder faithfully reconstructs the video. Given a cinematic video input $V$ and an encoding strategy $P$, the agentic encoder $E(\cdot)$ and fixed decoder $\mathrm{Dec}(\cdot)$ define:
\begin{equation}
    G = E(V;P), \qquad
    \hat V = \mathrm{Dec}(G).
    \label{eq:task-autoencoding}
\end{equation}

Here, the output of the encoder $G$ is a Film KG. It represents the input film through structured-text nodes describing cinematic content, typed edges encoding dependencies among these elements, and linked multimodal assets synthesized from the textual nodes (Sec.~\ref{sec:kg}). All source-video information used for reconstruction is mediated by the textual graph. Unlike dense neural embeddings, this explicit representation $G$ is designed to meet the three agent-native requirements in Sec.~\ref{sec:intro}. 

\subsection{Optimization Objective}
\label{sec:task-objective}

Under the fixed decoder, reconstruction fidelity measures how well $G$ preserves the source video. Let $R(V,\hat V)$ denote reconstruction discrepancies and $\mathcal L(\cdot)$ map them to a scalar loss:

\begin{equation}
    \ell(V,G)
    :=
    \mathcal L\!\left(R\!\left(V,\mathrm{Dec}(G)\right)\right).
    \label{eq:task-reconstruction-loss}
\end{equation}

From $G=E(V;P)$, this loss evaluates both (1) an \textbf{individual video representation} and (2) the \textbf{general encoding strategy} that produces it. We therefore formulate optimization at both the \textbf{strategy} and \textbf{instance} levels:
\begin{equation}
    \left\{
    \begin{aligned}
        &\min_{P\in\mathcal P}\;
        \mathbb E_{V'\sim\operatorname{Unif}(\mathcal D)}
        \left[\ell\!\left(V',E(V';P)\right)\right],
        &&\text{strategy level},\\[3pt]
        &\min_{G\in\mathcal G}\;
        \ell(V,G),
        &&\text{instance level}.
    \end{aligned}
    \right.
    \label{eq:objective}
\end{equation}

where $\mathcal D=\{V_i\}_{i=1}^{L}$ is the strategy-learning collection of $L$ videos. The strategy-level objective averages reconstruction loss over \(V'\sim\operatorname{Unif}(\mathcal D)\), while the instance-level objective minimizes it for a given \(V\). AVA-Encoder realizes these objectives through (1) \textbf{amortized encoding strategy learning}
and optional (2) \textbf{instance-specific Film KG refinement} within the dual-loop textual-gradient framework (Sec.~\ref{sec:dualloop}). Here, \emph{pseudo-training} fits $P$ to $\mathcal D$ from loss feedback and evaluates it on held-out videos as in training, but updates only the natural-language strategy, with model weights frozen.

\section{Method: AVA-Encoder}
\label{sec:method}

AVA-Encoder addresses the three challenges in Sec.~\ref{sec:intro} through three complementary designs (Figure~\ref{fig:pipeline}).

\textbf{(1)Representation: Agent-native Film KG.} To preserve cinematic detail and its dependencies, a hierarchical encoder captures film-, shot-, and keyframe-level information in editable text nodes linked by typed edges across modalities and entities (Secs.~\ref{sec:encoder}--\ref{sec:kg}).

\textbf{(2)Architecture: Agentic auto-encoding paradigm.} In the absence of ground-truth Film KGs, we use the source video itself as indirect supervision for learning its representation. As shown in Figure~\ref{fig:pipeline}(a), the agent encodes the source video into a Film KG, a fixed decoder reconstructs the video from the graph, and the agent compares the reconstruction with the source at the shot and keyframe levels. This comparison grounds the optimization signal in verifiable source facts and identifies discrepancies that guide revisions to the encoding strategy or the graph itself, closing the loop for reconstruction-driven agentic self-evolution.

\textbf{(3)Learning: Dual-loop self-evolution.} To reduce test-time search and limit overfitting to individual films, the outer loop uses this feedback across videos to learn a reusable encoding strategy. The optional inner loop then refines the graph of a new film where that strategy still misses specific details (Sec.~\ref{sec:dualloop}).

\subsection{Agentic Video Encoder}
\label{sec:encoder}

Mapping a high-dimensional film into structured text \textbf{risks losing fine-grained cinematic information} (Sec.~\ref{sec:intro}). The Agentic Video Encoder addresses this through \emph{three-level specialized understanding} across film, shot, and keyframe granularities, together with \emph{cross-level context injection} to preserve dependencies across these levels (Figure~\ref{fig:pipeline}(b)).

Given a video $V$ partitioned into ordered shots and sampled keyframes, the encoder utilizes a hierarchical strategy $P$, composed of three level-specific system prompts $(P_{\mathrm{film}}, P_{\mathrm{shot}}, P_{\mathrm{kf}})$, to guide the extraction process. Crucially, upper-level extracted information is injected into finer levels as context to ensure consistency:

\begin{itemize}[nosep,leftmargin=1.5em]
    \item \textbf{Film-level} captures the global narrative, cross-shot dependencies and builds a shared registry $\mathcal B_{\mathrm{reg}}$ of canonical entities and film-wide identifiers.
    \item \textbf{Shot-level} captures temporal, audiovisual, and camera dynamics, explicitly conditioned on the film-level context and the shared registry $\mathcal B_{\mathrm{reg}}$.
    \item \textbf{Keyframe-level} extracts fine-grained visual composition and appearance, guided by its parent shot's description and the same registry $\mathcal B_{\mathrm{reg}}$.
\end{itemize}

Finally, a deterministic function maps these extracted structured descriptions and relations into the nodes and typed edges of the Film KG $G=E(V;P)$, requiring no additional model calls. The encoding strategy $P$ is optimized through outer-loop Amortized Encoding Strategy Learning (Sec.~\ref{sec:outer-loop}).

\subsection{Film KG Representation}
\label{sec:kg}
\paragraph{KG Representation Structure.}
To preserve the cinematic multimodal dependencies(Sec.~\ref{sec:intro}) in an agent-native form, we construct the \textbf{Film KG}:
$G=(\mathcal N_G,\mathcal E_G,\mathcal A_G)$.
Here, $\mathcal N_G$ contains structured-text nodes storing cinematic contents, $\mathcal E_G$ contains typed edges capturing their dependencies, and $\mathcal A_G$ stores linked generated multimodal assets. 

First, to make multimodal content individually accessible and editable by agents, we decouple the film into separately addressable \textbf{textual nodes} of types $\{$Story, Event, Shot, Character, Scene, Object, Camera, Style, Audio, Keyframe$\}$. 

Second, \textbf{typed edges} explicitly map out structural, temporal, and semantic dependencies among nodes, enabling coordinated editing and connecting textual edits directly to \textbf{multimodal asset} regeneration.

\paragraph{KG Topological Editting.} When a node is edited, we determine the full scope of affected nodes through an edit-conditioned, three-stage dependency traversal:
\begin{enumerate}
\item \textbf{Material Propagation:} Starting from the edited nodes, we repeatedly follow the material dependency edges allowed for this edit type until no new nodes are reached.
\item \textbf{Semantic Consistency:} We then add semantic dependents that require consistency checks (e.g., verifying if the dialogue still fits a newly edited character).
\item \textbf{Hierarchical Update:} Finally, we update the enclosing \emph{event} and \emph{story} descriptions so that the global narrative accurately reflects the lower-level edits.
\end{enumerate}
(See Appendix for detailed editing topologies.)

\subsection{Dual-Loop Self-Evolution}
\label{sec:dualloop}

\begin{figure}[tbp]
    \centering
    \begin{minipage}[c]{0.5\linewidth}
        \centering
        \includegraphics[height=7.5cm]{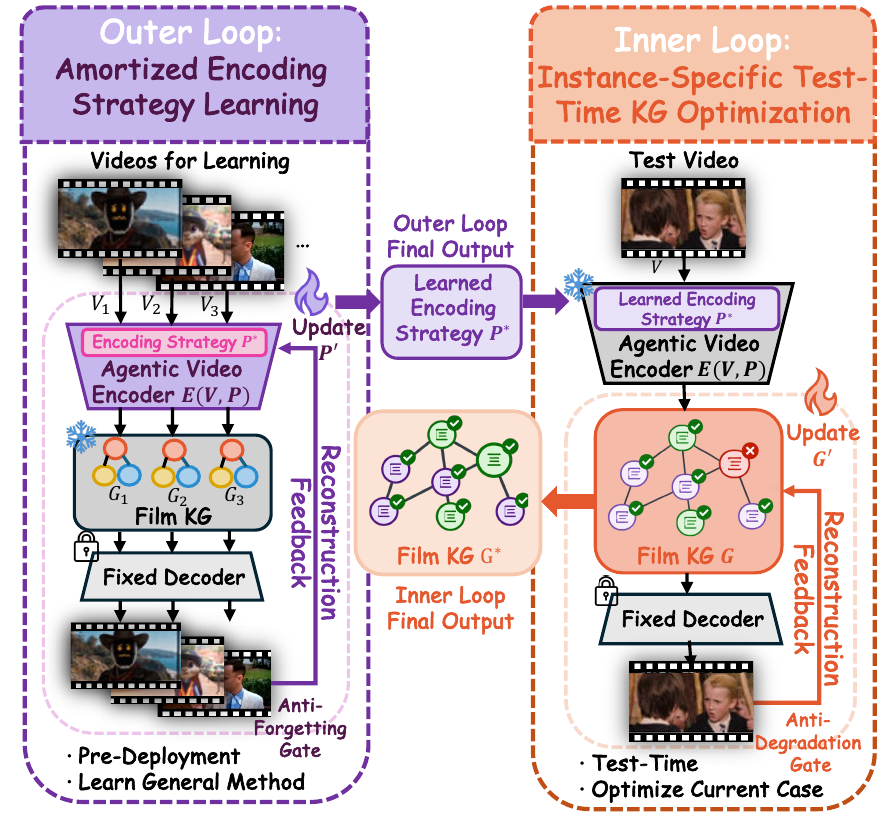}
        \captionof{figure}{\textbf{Dual-loop intuition.} The outer loop amortizes encoding strategy learning across multiple videos. The inner loop performs instance-specific test-time KG optimization.}
        \label{fig:dual_loop_intuition}
    \end{minipage}\hfill
    \begin{minipage}[c]{0.48\linewidth}
        \centering
        \includegraphics[height=11cm]{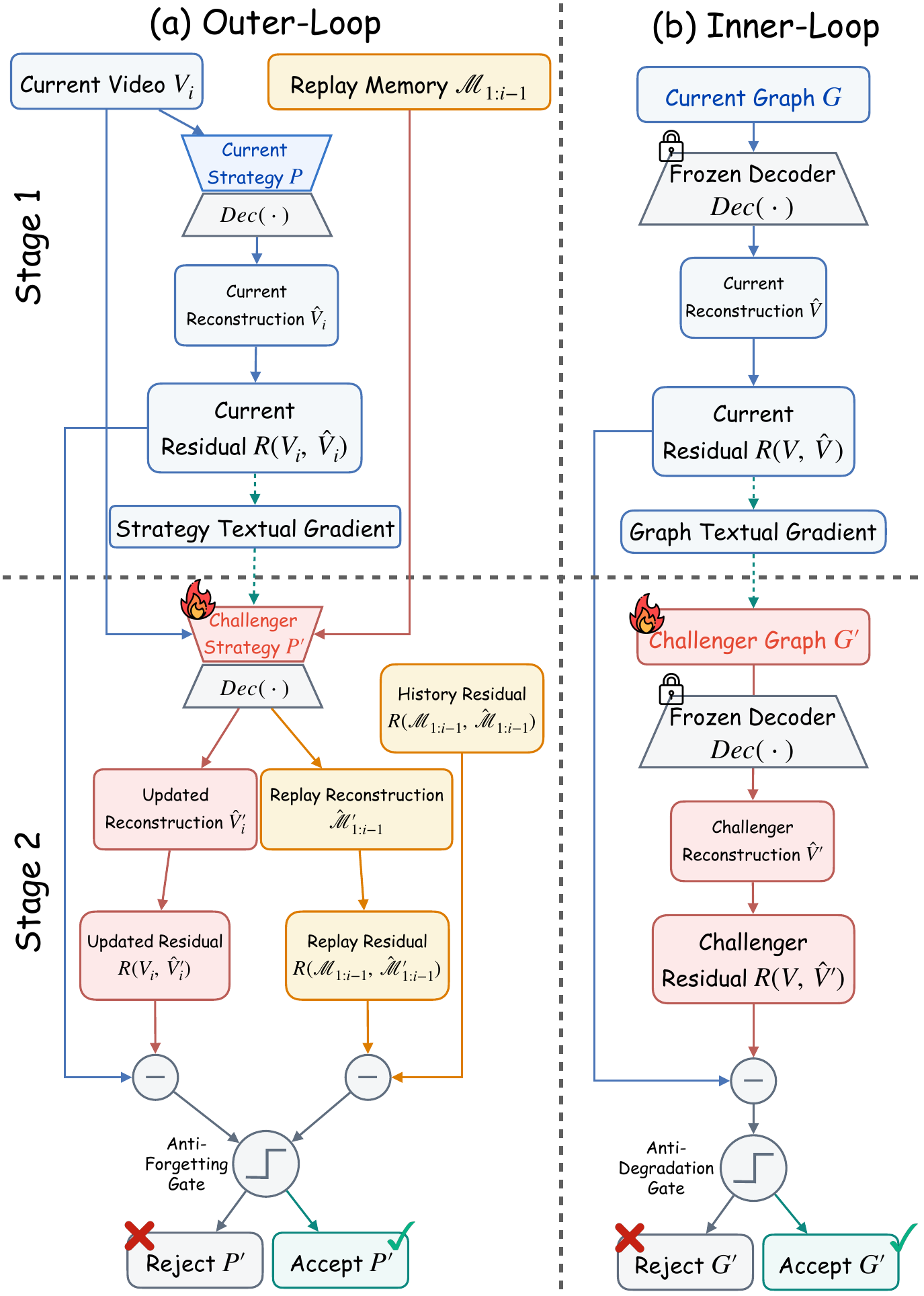}
        \captionof{figure}{\textbf{Mechanics of the acceptance gates.} (a) Anti-Forgetting Gate (outer loop) and (b) Anti-Degradation Gate (inner loop).}
        \label{fig:dual-loop}
    \end{minipage}
\end{figure}

\begin{algorithm}[tbp]
\caption{Dual-Loop Textual-Gradient Evolution}
\label{alg:unified_dual_loop}
\AlgStyle
\begin{algorithmic}[1]
\vspace{2pt}
\Statex \textbf{Phase 1: Outer Loop (Amortized Encoding Strategy Learning)}
\Statex \textbf{Input:} Training videos $\{V_i\}_{i=1}^L$, initial strategy $P_0$
\State $P \gets P_0$
\For{each $V_i$}
    \State Encode $G_i \gets E(V_i; P)$ and decode $\hat{V}_i \gets \operatorname{Dec}(G_i)$
    \State Assess $\hat{V}_i$ against $V_i$ to get reward $R_{\text{reward}}$ and errors $\mathcal{F}_P$
    \State Propose $P' \gets \operatorname{Propose}_P(P, \nabla_{\text{text}}(P, \mathcal{F}_P))$
    \If{$P'$ improves $R_{\text{reward}}$ and passes Anti-Forgetting Gate} \label{line:outer-gate-main} \Comment{Check generalization}
        \State $P \gets P'$
    \EndIf
\EndFor
\State \textbf{Output:} Learned general strategy $P^*$

\vspace{6pt}
\hrule
\vspace{6pt}

\Statex \textbf{Phase 2: Inner Loop (Instance-Specific Test-Time KG Optimization)}
\Statex \textbf{Input:} Test video $V_{\text{test}}$, fixed strategy $P^*$
\State $G \gets E(V_{\text{test}}; P^*)$ \Comment{Fast single-pass encoding}
\For{$t = 1 \dots T_{\text{inner}}$}
    \State Decode $\hat{V}_{\text{test}} \gets \operatorname{Dec}(G)$
    \State Assess $\hat{V}_{\text{test}}$ against $V_{\text{test}}$ to get reward $R_{\text{reward}}$ and localized errors $\mathcal{F}_G$
    \State \textbf{if} no errors found \textbf{then break}
    \State Propose $G' \gets \operatorname{Propose}_G(G, \nabla_{\text{text}}(G, \mathcal{F}_G))$
    \If{$G'$ improves $R_{\text{reward}}$ and passes Anti-Degradation Gate} \label{line:inner-gate-main} \Comment{Check local fidelity}
        \State $G \gets G'$
    \EndIf
\EndFor
\State \textbf{Output:} Refined specific Film KG $G^*$
\end{algorithmic}
\end{algorithm}

Relying purely on test-time search to optimize the graph is computationally prohibitive and prone to overfitting to specific instances (Sec.~\ref{sec:intro} Challenge 3). To balance inference overhead and enable generalization, we formulate agentic self-evolution~\cite{autodreamer} as a \textbf{dual-loop optimization architecture} (Algorithm~\ref{alg:unified_dual_loop}):

\begin{itemize}[nosep,leftmargin=1.5em]
    \item \textbf{(1) Outer Loop (Amortized Strategy Learning).} Before deployment, the outer loop amortizes the cost of textual-gradient search across a set of learning videos by revising a shared encoding strategy $P$. The learned strategy $P^*$ then encodes an unseen film into an initial Film KG $G$ in a single pass (Figure~\ref{fig:dual_loop_intuition} and Algorithm~\ref{alg:unified_dual_loop}).
    \item \textbf{(2) Inner Loop (Instance-Specific KG Optimization).} At test time, even a well-learned strategy may miss details specific to a film. When finer reconstruction is needed, the optional inner loop keeps $P^*$ fixed and uses reconstruction feedback to refine that film's graph $G$. This reserves additional search for details that the shared strategy did not capture on its first pass.
\end{itemize}

To operationalize this architecture, we first define the optimization signals used for evaluation and critique (Sec.~\ref{sec:reconstruction-error}), followed by the specific mechanics of the outer and inner loops (Secs.~\ref{sec:outer-loop} and~\ref{sec:inner-loop}). In each loop, the reconstructions of the current solution and of the challenger are compared against the source video, and the gate decides whether the challenger is accepted (Figure~\ref{fig:dual-loop}).

\subsubsection{Optimization Signal and Textual Feedback}
\label{sec:reconstruction-error}

Evaluating open-ended visual creations directly with a vision-language model (VLM) is unreliable for iterative optimization~\cite{badseeing}. (1) First, holistic VLM scores are unstable: the same output can receive different ratings across rounds, making it difficult to judge whether a revision truly helped. (2) Second, overall scores easily reflect subjective model preferences regarding visual style or composition. (3) Third, a single scalar score is hard to diagnose, as it does not tell the agent which shot lost what information or which part of the representation should be revised. Although no ground-truth Film KG exists, we can turn this into a verifiable comparison by decoding $G$ into a reconstructed video $\hat{V}$ with a fixed decoder and using the source video $V$ as a shared reference anchor to check which original facts are preserved in each round.

To obtain both reliable scoring and actionable revision guidance from the same reconstruction evidence, we decompose evaluation into atomic factual checks at the shot and keyframe levels~\cite{sotopiarl,rare}. (1) First, for each source shot and keyframe, we generate a fixed set of factual QA pairs once from the source video and reuse the exact same questions in every subsequent round. (2) Second, a VLM verifies each observable fact in $\hat{V}$, reducing subjective bias; the fraction of preserved facts defines the optimization reward $R_{\mathrm{reward}}$, making candidate reconstructions directly comparable before and after a revision. (3) Third, every failed QA item retains natural-language evidence contrasting what appears in the source video with what was reconstructed~\cite{rationalrewards}. Following TextGrad~\cite{textgrad}, we distill these discrepancies into targeted natural-language critiques (textual gradients) to update either the encoding strategy $P$ or the graph $G$. Both loops are supervised solely by this reconstruction consistency signal (see Appendix for details), while a separate metric, $R_{\mathrm{eval}}$, is strictly held out for final evaluation (Sec.~\ref{sec:exp-setup}).

\subsubsection{Outer Loop: Amortized Encoding Strategy Learning}
\label{sec:outer-loop}

Relying on manually crafted prompts or per-video test-time search alone is neither scalable nor generalizable. On the one hand, different films rely on distinct narrative structures, camera languages, and audiovisual cues; hand-crafting a prompt that covers these variations while remaining concise is difficult, and it is rarely clear which added prompt tokens actually improve encoding. On the other hand, running repeated generation, evaluation, and graph editing on every unseen video incurs high test-time cost, and lessons learned from fixing one shot do not carry over to future films. This raises a natural question: starting from a concise initial prompt $P_0$, can an agent autonomously learn a reusable hierarchical encoding strategy solely from reconstruction consistency, so that new videos benefit on their very first encoding pass?

The outer loop answers this by discovering and filtering general encoding rules across a small collection of learning videos before deployment (Figure~\ref{fig:dual_loop_intuition}; Algorithm~\ref{alg:unified_dual_loop}). On each video, the agent inspects the failure evidence from $R_{\mathrm{reward}}$ and revises the shared strategy $P$ into a candidate $P'$; for example, generalizing a single camera-timing error into a rule that explicitly records the temporal order between camera movement and character action. We then re-encode and reconstruct with $P'$. To prevent the strategy from overfitting to the current video, the \textbf{Anti-Forgetting Gate} (Algorithm~\ref{alg:unified_dual_loop}, Line~\ref{line:outer-gate-main}; Figure~\ref{fig:dual-loop}a) accepts $P'$ only if it improves $R_{\mathrm{reward}}$ on the current video while keeping any overall regression on historical replay samples, drawn across all previously seen learning videos, within a fixed tolerance. Each accepted strategy is carried forward to the next learning video, eventually yielding a general strategy $P^*$ whose pre-deployment search cost is amortized over all subsequent films. To control prompt redundancy during evolution, we also cap the total prompt length and encourage merging or replacing existing instructions rather than stacking new ones (details on replay sampling, gate thresholds, and length constraints are in Appendix).

\subsubsection{Inner Loop: Instance-Specific Test-Time KG Optimization}
\label{sec:inner-loop}

Even a well-generalized strategy $P^*$ cannot eliminate every instance-specific error on unseen films. (1)First, general rules still miss fine-grained local facts: a strategy that knows to attend to the relation between character action and camera movement may still invert their order in a particular shot. (2)Second, local edits to a textual graph can have unstable side effects, because the decoder re-renders multiple coupled visual attributes from the updated text; fixing camera timing in a node may inadvertently break a previously accurate character position. Effective instance-level refinement therefore requires both localizing residual errors on the current graph and verifying that each edit does not harm other attributes in the same reconstruction.

To handle these residual errors, the inner loop builds directly on the outer loop (Figure~\ref{fig:dual_loop_intuition}; Algorithm~\ref{alg:unified_dual_loop}). Given an unseen test video $V_{\mathrm{test}}$, the encoder first generates an initial graph $G = E(V_{\mathrm{test}}; P^*)$ in a single pass, and invokes test-time refinement only where needed. Keeping $P^*$ fixed, the agent uses the failed QA items and their textual evidence to pinpoint the responsible shot or keyframe nodes, proposes a revised graph $G'$, and re-renders only the affected content. The \textbf{Anti-Degradation Gate} (Algorithm~\ref{alg:unified_dual_loop}, Line~\ref{line:inner-gate-main}; Figure~\ref{fig:dual-loop}b) then verifies net improvement across dimensions: it accepts $G'$ only if the overall reward $R_{\mathrm{reward}}$ of that shot or keyframe increases while total regression across non-target dimensions stays within a fixed tolerance, iteratively producing the refined graph $G^*$ (see Appendix). Taken together, the two loops play complementary roles: the outer loop distills reusable encoding rules into $P^*$ and validates them \emph{across videos}, whereas the inner loop fixes instance-specific details in $G^*$ and validates them \emph{across dimensions} within the current video, confining extra test-time compute strictly to content that still requires refinement.

\section{Experiments}
\label{sec:experiments}

We first establish the experimental setup, followed by our main results demonstrating end-to-end video reconstruction fidelity and the cumulative contributions of our architectural designs. Finally, we provide a dedicated discussion section centered around three Research Questions (RQs) to deeply analyze the representation efficiency, self-evolution generalization, and downstream utility of our framework.

\begin{figure}[t]
    \centering
    \begin{minipage}[c]{0.48\linewidth}
        \centering
        \includegraphics[width=\linewidth]{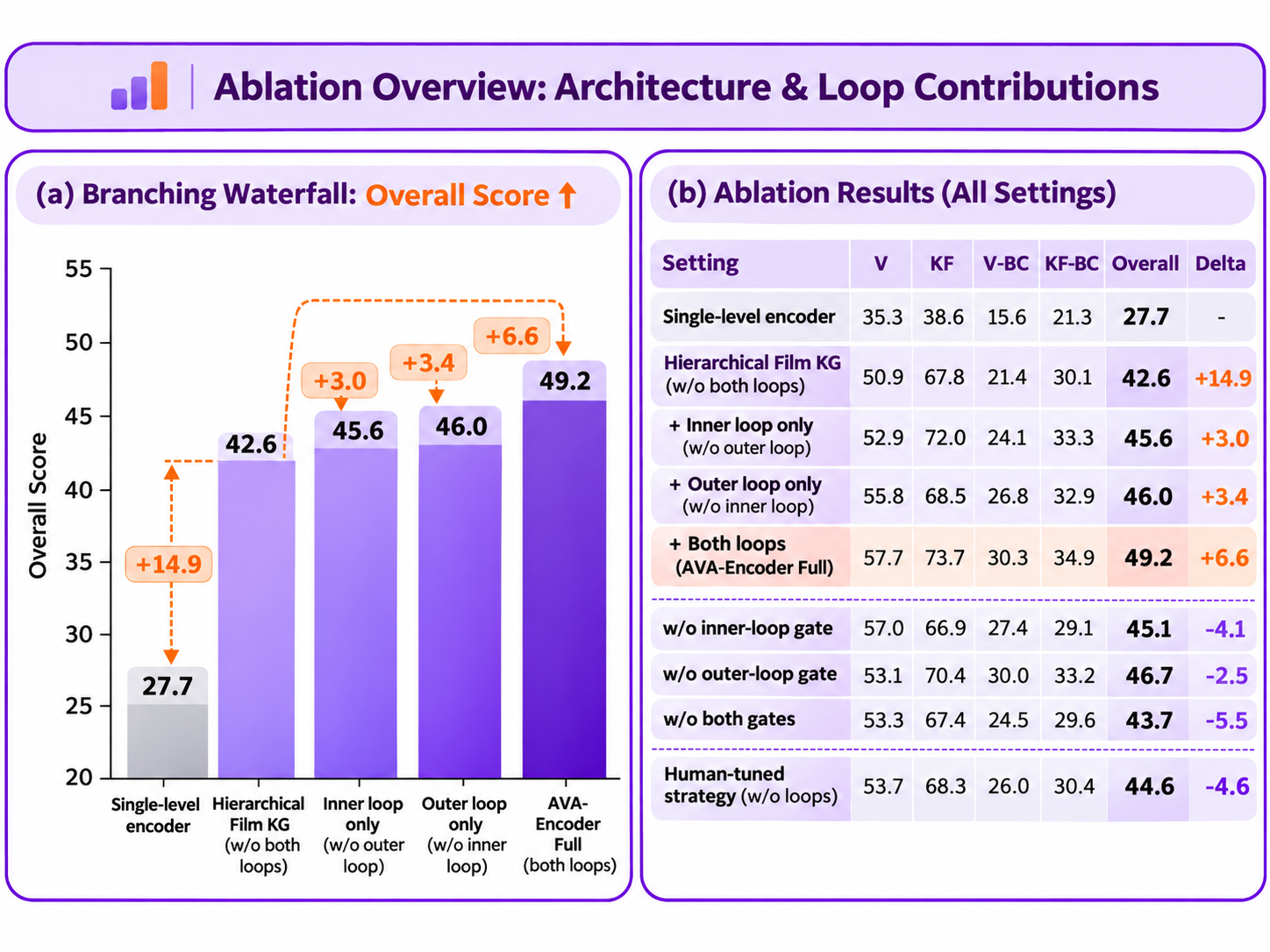}
        \captionof{figure}{\textbf{Ablation overview} (\%; higher is better). (a) Overall-fidelity build-up from the single-level encoder to the full model. (b) All ablation settings, including gate removals and the human-tuned strategy.}
        \label{fig:ablation}
    \end{minipage}\hfill
    \begin{minipage}[c]{0.50\linewidth}
        \centering
        \includegraphics[width=\linewidth]{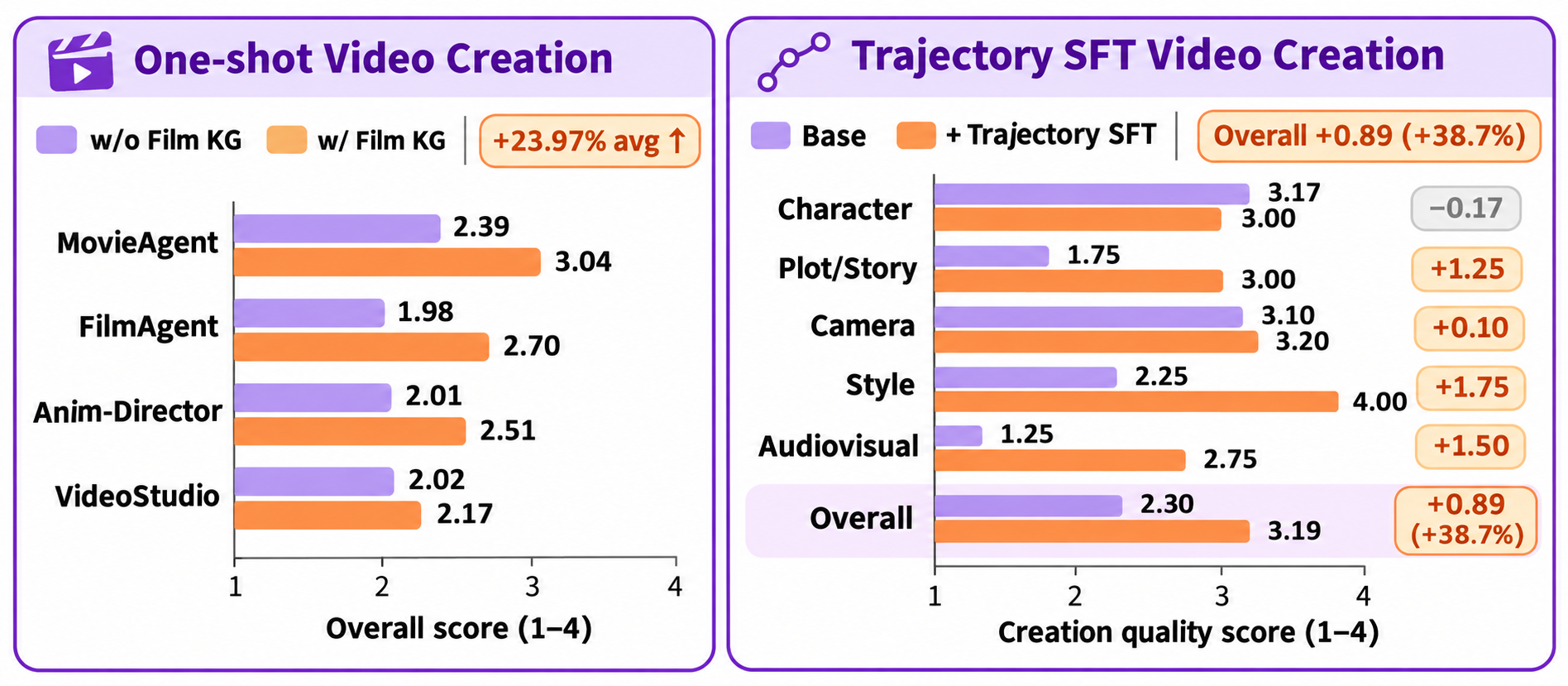}
        \captionof{figure}{\textbf{Downstream utility} (higher is better). Left: one-shot video creation with four open-source frameworks, without vs.\ with the Film KG supplied as text. Right: Qwen3-8B before vs.\ after trajectory SFT on 7,863 reverse-engineered trajectories.}
        \label{fig:downstream}
    \end{minipage}
\end{figure}

\subsection{Experimental Setup}
\label{sec:exp-setup}

\paragraph{Data.}
The agent learns general encoding strategy on 6 videos, and evaluates on 23 non-overlapping videos spanning animation, human-directed AI short films, classic cinema, and newly released films to test generalization. See Appendix Sec.~\ref{app:reproducibility} for details.

\paragraph{Models and comparison protocol.}

Qwen-3.7-Plus~\cite{qwen37plus} performs hierarchical understanding within the Agentic Video Encoder and revises the policy or Film KG using textual gradients. Gemini-3.1-Pro-Preview~\cite{gemini31pro} performs factual judgments for $R_{\mathrm{reward}}$, and Qwen-3.8-Max~\cite{qwen38max} for $R_{\mathrm{eval}}$. Nano Banana Pro~\cite{nanobananapro} and HappyHorse 1.0~\cite{happyhorse10} generate decoder images and videos, respectively. All model weights remain fixed.

VideoAnalyzer~\cite{videoanalyzer}, Storyboard Studio~\cite{storyboardstudio}, soap2soap~\cite{soap2soap}, and AVA-Encoder share the same video understanding model, test videos, generation backbones, and decoder-interface limits (5 generated reference keyframes and 1,200 shot-prompt tokens). Reconstruction generators never receive source-video pixels. Appendix Secs.~\ref{app:complete-prompts} and~\ref{app:baselines} provide prompts and detailed budget comparisons.

\paragraph{Final reconstruction evaluation.}
We use $R_{\mathrm{eval}}$ to compare how well different representations preserve source content under the shared reconstruction protocol. This evaluation is reserved for reporting, and its scores do not enter either optimization loop.

\textbf{(1) Four complementary comparisons.} Direct Video (V) and direct Keyframe (KF) compare reconstructed shots and keyframes with their source counterparts in the original media. Video back-captioning (V-BC) and Keyframe back-captioning (KF-BC) describe the source and the reconstruction independently, then compare the two descriptions in text space. Each caption is produced without access to its counterpart.

\textbf{(2) Source-referenced judgments.} Each direction checks whether observable source content is preserved along the applicable cinematic dimensions~\cite{pymeta}: Character, Scene, Position, Motion, Audio, Style, Camera, and Narrative. Shot-level comparisons can assess temporal and audio evidence, while keyframe comparisons are limited to what can be observed in a static image.

\textbf{(3) Aggregation.} We average dimension scores within each shot or keyframe, aggregate those scores within each video, and then average across videos for each direction. Overall is the mean of the four direction scores. Appendix provides the full scoring rules and human-agreement results.

\subsection{Main Results}
\label{sec:exp-main}

\begin{figure}[!tbp]
    \centering
    \includegraphics[width=\linewidth,trim=70 0 0 0,clip]{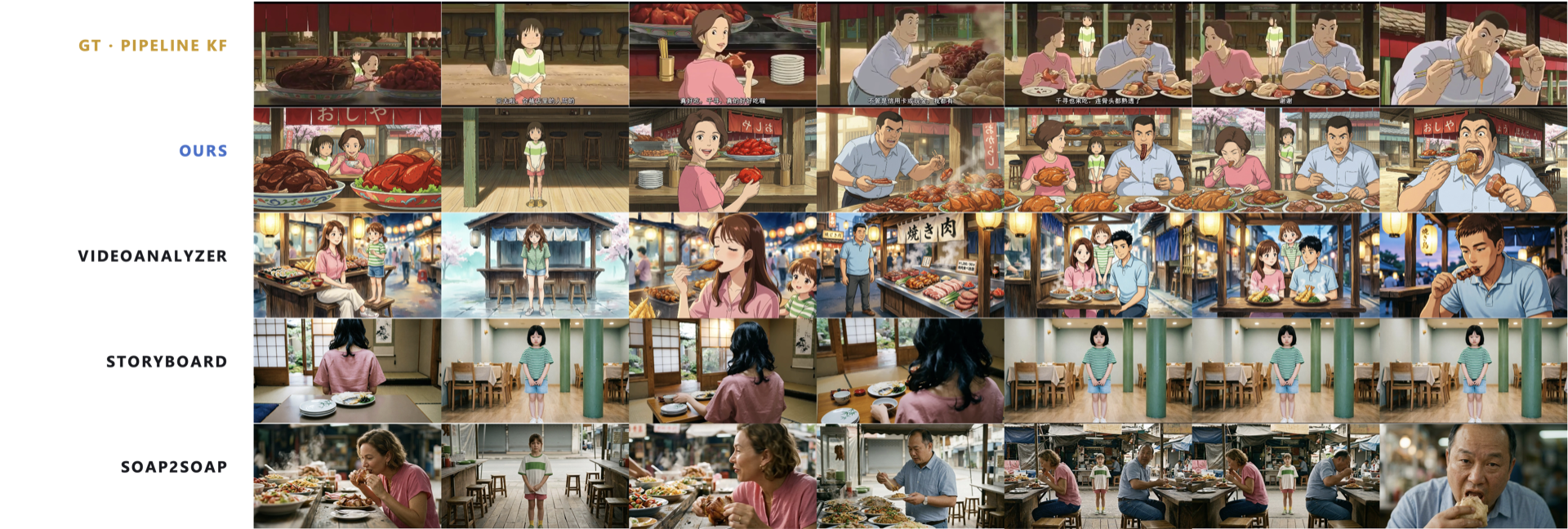}
    \caption{Reconstruction comparison between AVA-Encoder and baselines, showing the preservation of character identity, visual detail, and visual consistency.}
    \label{fig:reconstruct_compare}
\end{figure}

\paragraph{Reconstruction fidelity.}
Under the shared decoder and interface budget, AVA-Encoder achieves 49.2\% Overall fidelity, surpassing the strongest baseline, soap2soap, by 20.9 percentage points and leading across all four evaluation directions (Table~\ref{tab:main}). Gains in both direct and caption-based comparisons demonstrate our representation's superior ability to preserve perceptual details and semantic dependencies. Figure~\ref{fig:reconstruct_compare} shows a representative comparison; Appendix provides further examples.

\begin{table}[t]
    \centering
    \small
    \setlength{\tabcolsep}{6pt}
    \renewcommand{\arraystretch}{1.06}
    \caption{\textbf{Reconstruction fidelity} (\%; higher is better). V/KF: direct video/keyframe comparison; V-BC/KF-BC: video/keyframe back-captioning. Overall averages the four scores.}
    \label{tab:main}
    \begin{tabularx}{0.62\linewidth}{@{}>{\raggedright\arraybackslash}Xccccc@{}}
        \toprule
        \textbf{Method} & V & KF & V-BC & KF-BC & \textbf{Overall} \\
        \midrule
        VideoAnalyzer & 26.1 & 28.5 & 9.7 & 21.7 & 21.5 \\
        Storyboard Studio & 16.4 & 28.6 & 9.6 & 23.0 & 19.4 \\
        soap2soap & 36.7 & 39.5 & 15.8 & 21.3 & 28.3 \\
        \midrule
        \rowcolor{tablehl}
        \textbf{AVA-Encoder (ours)} & \textbf{57.7} & \textbf{73.7} & \textbf{30.3} & \textbf{34.9} & \textbf{49.2} \\
        \bottomrule
        \end{tabularx}
\end{table}

\subsection{Discussion}
\label{sec:exp-discussion}
\label{sec:exp-ablation}
\label{sec:exp-applications}

\begin{figure}[!tbp]
    \centering
    \begin{subfigure}[t]{\linewidth}
        \centering
        \includegraphics[width=0.9\linewidth]{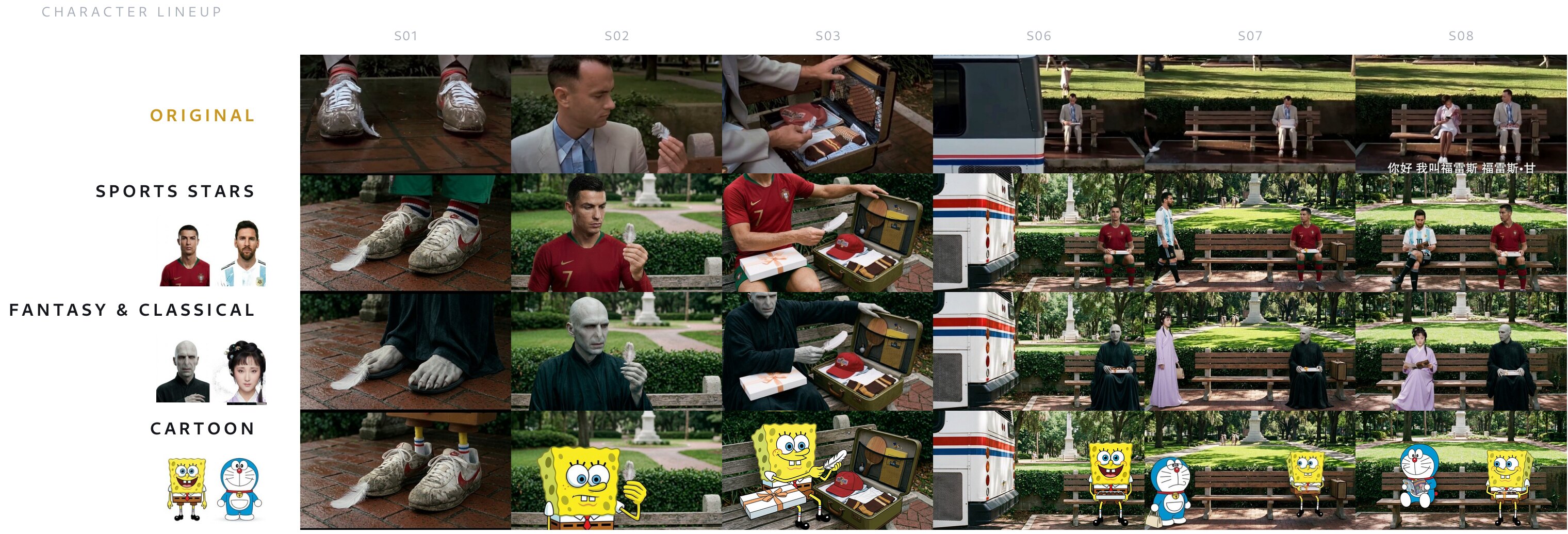}
        \caption{Character replacement across a six-shot sequence. A single character edit jointly updates the dependent keyframes and shots.}
        \label{fig:qualitative-identity}
    \end{subfigure}\\[4pt]
    \begin{subfigure}[t]{\linewidth}
        \centering
        \includegraphics[width=0.9\linewidth]{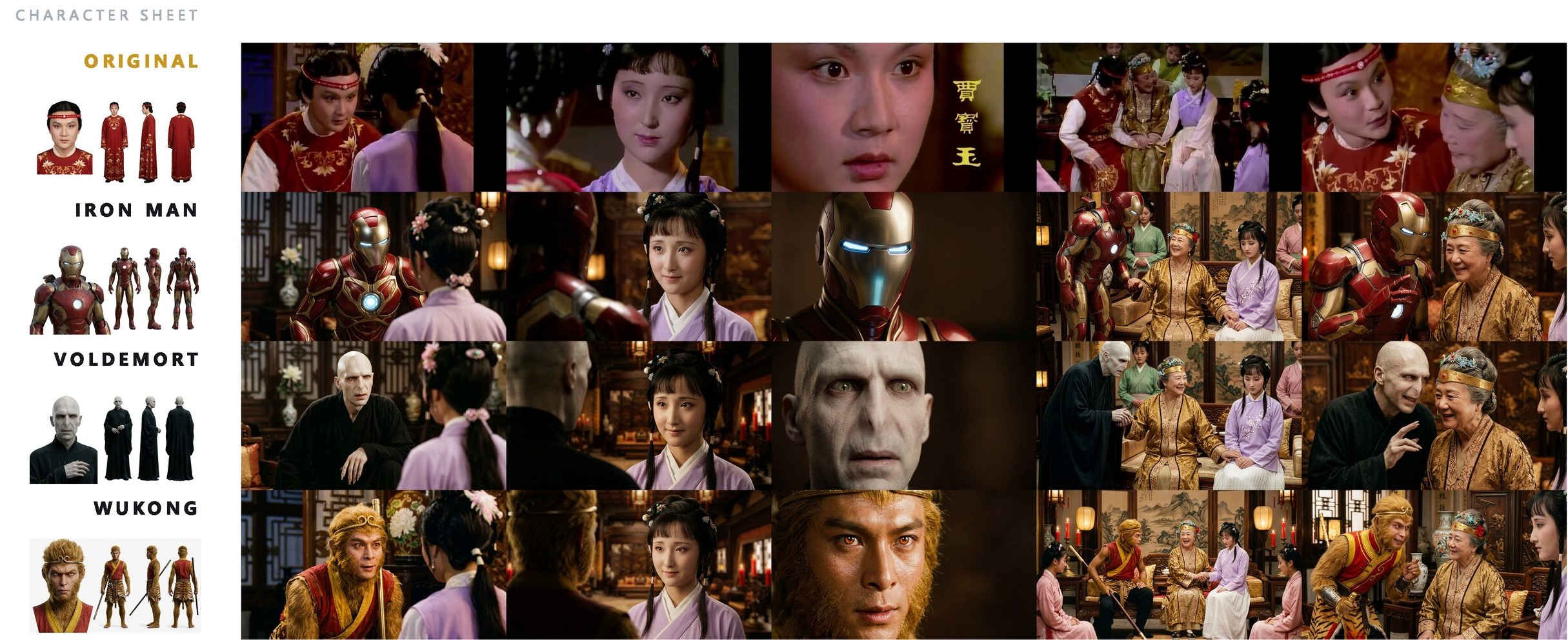}
        \caption{Cross-domain coordinated identity replacement in a film and TV-drama clip. The edit propagates through the character's state transitions to every linked keyframe and shot. For recognizability, this editing demonstration explicitly requests named characters; it is not part of the released dataset.}
        \label{fig:qualitative-identity-classical}
    \end{subfigure}\\[4pt]
    \begin{subfigure}[t]{\linewidth}
        \centering
        \includegraphics[width=0.9\linewidth]{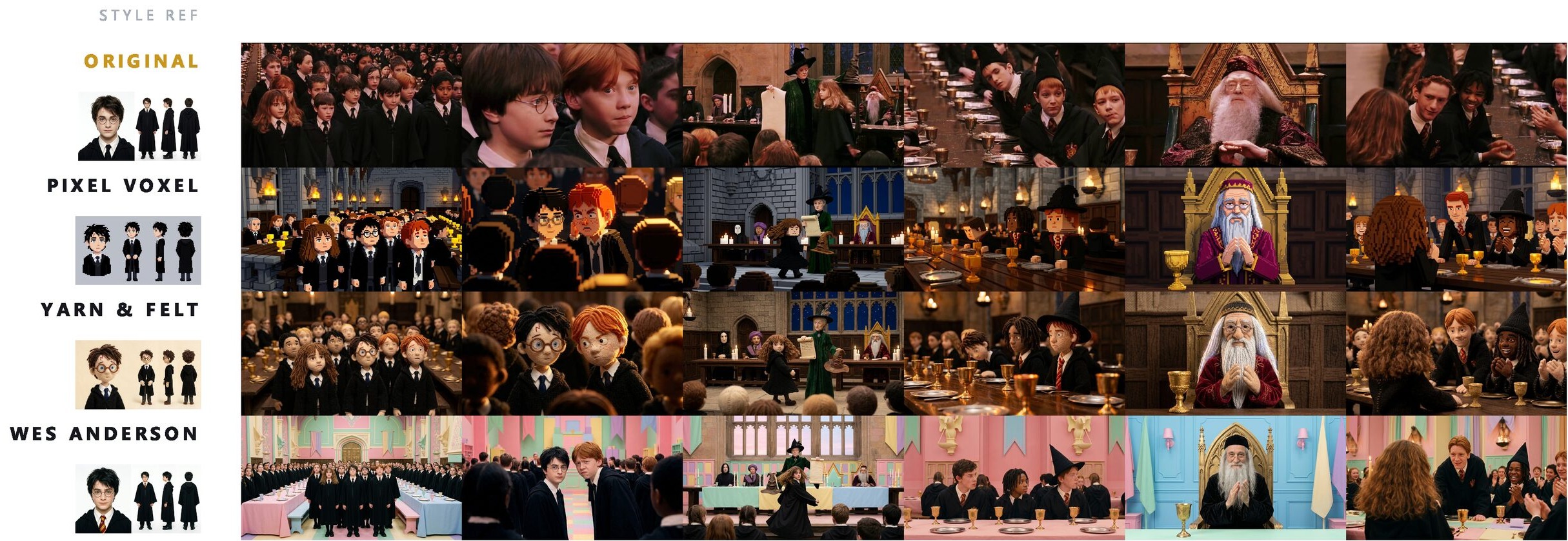}
        \caption{Visual-treatment editing. The edit follows style-state transitions and shot bindings, applying one consistent treatment to the linked sequence without rewriting each shot independently.}
        \label{fig:qualitative-style}
    \end{subfigure}
    \caption{\textbf{Editing showcases.} Coordinated graph-based editing with the Film KG.}
    \label{fig:editing-showcases}
\end{figure}

\paragraph{RQ1: Representation vs. Verbosity.}
\emph{Does the fidelity improvement stem from the graph structure, or simply from generating longer text descriptions?} To isolate this, we established a ``Rich Descriptions'' control that covered the same eight dimensions but lacked explicit hierarchical or topological structure. As detailed in Appendix Sec.~\ref{app:ablations}, these unstructured descriptions required over 3,400 prompt tokens per shot but only achieved 37.9\% Overall fidelity. In contrast, the Agent-Native Film KG structure achieved 50.7\% while using 88\% fewer tokens (410.7 tokens per shot). This confirms that faithful reconstruction relies heavily on carefully engineered structural dependencies, rather than mere verbosity.

\paragraph{RQ2: Self-Evolution and Generalization.}
\emph{Can the agent autonomously learn a generalizable representation strategy purely from reconstruction feedback?} Figure~\ref{fig:ablation} isolates the independent and combined contributions of our encoding structure, dual-loop optimization, and acceptance gates.

\paragraph{(1)Hierarchical encoding and KG representation provides the foundation.}
First, even without any test-time optimization loops, introducing the hierarchical Film KG structure raises the Overall fidelity by 14.9 points over the single-level encoder (from 27.7\% to 42.6\%). This decisively validates that multi-level understanding is critical for faithful video representation.

\paragraph{(2)Dual loops complement each other.}
Building upon the hierarchical KG, we isolate the impact of each loop. Activating only the inner loop to refine instance-specific details yields a +3.0 point gain. Activating only the outer loop to learn a shared, amortized strategy yields a +3.4 point gain. When combined, the dual loops achieve a cumulative +6.6 point improvement over the initial hierarchical configuration, reaching 49.2\%. Notably, the purely self-evolved strategy from the outer loop (46.0\%) outperforms a meticulously human-tuned strategy (44.6\%), while utilizing over 70\% fewer system-prompt tokens. 

\paragraph{(3)Gates ensure stable evolution and better generalization.}
Finally, targeted ablations confirm that unrestricted evolution leads to representation collapse. Removing the Anti-Degradation Gate (inner) or the Anti-Forgetting Gate (outer) lowers Overall fidelity by 4.1 and 2.5 points from the full model, respectively. Removing both gates drops performance drastically to 43.7\% (-5.5 points), confirming that strict acceptance control is necessary to ensure generalization performance during autonomous learning.

\paragraph{RQ3: Downstream Utility.}
\emph{Do these KGs and reverse-engineered trajectories possess practical utility for downstream video generation?} We evaluate them across three tasks:

1. \emph{Trajectory SFT for creation:} High-quality planning trajectories are critically scarce. We sampled 145 videos and 7,863 agentic video creation trajectories from our dataset to supervise fine-tune (SFT) Qwen3-8B. Evaluated on 5 independent test cases, this trajectory SFT improved Overall video creation quality by \textbf{38.7\%} (from 2.30 to 3.19; Figure~\ref{fig:downstream}, right), directly validating the training value of our reverse-engineered data.

2. \emph{One-shot downstream generation:} We provided each of four open-source video-creation frameworks with the complete Film KG as text and a one-sentence usage request. As shown in Figure~\ref{fig:downstream} (left), Overall quality improves across all tested frameworks without any workflow modifications or adapter training, demonstrating cross-system compatibility via a simple textual interface.

3. \emph{Graph-based editing:} The Film KG supports targeted edits with coordinated updates to dependent content. Figure~\ref{fig:qualitative-identity} shows consistent character replacement across six linked shots, preserving unrelated characters, scenes, and event structures (see also Figures~\ref{fig:qualitative-identity-classical} and~\ref{fig:qualitative-style}). Appendix Sec.~\ref{app:topology-edit} provides further examples.

\section{Conclusion}
\label{sec:conclusion}

We present AVA-Encoder, an agentic video auto-encoding framework that transforms cinematic videos into structured, agent-native Film Knowledge Graphs (Film KGs). To overcome the absence of ground-truth graphs and expert creation trajectories, we introduce a reconstruction-driven dual-loop self-evolution mechanism. This architecture balances inference overhead with generalization by coupling an amortized outer loop for shared strategy learning with a targeted inner loop for instance-specific refinement. AVA-Encoder achieves 49.2\% Overall reconstruction fidelity, outperforming the strongest baseline by 20.9 percentage points. Furthermore, the resulting Film KGs and trajectories drive practical improvements in coordinated video editing, one-shot video generation, and trajectory-based agent fine-tuning. These results demonstrate a clear path toward bridging video and agent spaces, making the creation knowledge embedded in finished films explicit, editable, and reusable by autonomous agents.

\begingroup
\setlength{\emergencystretch}{3em}
\bibliography{conference}
\bibliographystyle{conference}
\endgroup

\appendix
\clearpage

\pdfbookmark[0]{Appendix}{appendix-title}
{\centering\Large\bfseries Appendix\par}
\vspace{1em}
\raggedbottom

\paragraph{Organization.}
This appendix expands the method (Sec.~\ref{sec:method}) and the experiments (Sec.~\ref{sec:experiments}) of the main paper. An unnumbered overview first summarizes how the framework and its components fit together, and Table~\ref{tab:app-organization} lists each lettered section and the part of the main paper it supports.

\begin{table}[h]
\centering
\footnotesize
\setlength{\tabcolsep}{3pt}
\caption{Organization of the appendix.}
\label{tab:app-organization}
\begin{tabular}{@{}lp{0.80\textwidth}@{}}
\toprule
Sec. & Content and supported main-paper part \\
\midrule
\ref{app:textual-gradients} & Dual-loop self-evolution: gates, algorithms, textual gradients (Sec.~\ref{sec:dualloop}) \\
\ref{app:evaluator} & Final evaluation $R_{\mathrm{eval}}$ and human alignment (Sec.~\ref{sec:exp-setup}) \\
\ref{app:qa-reward} & Optimization reward $R_{\mathrm{reward}}$ (Sec.~\ref{sec:reconstruction-error}) \\
\ref{app:dataset} & Released Film KG dataset and creation trajectories \\
\ref{app:qualitative} & Additional reconstruction comparisons (Sec.~\ref{sec:exp-main}) \\
\ref{app:topology-edit} & Graph-based editing visualizations (Sec.~\ref{sec:exp-discussion}) \\
\ref{app:baselines} & Reconstruction inference budget and baseline protocol (Sec.~\ref{sec:exp-setup}) \\
\ref{app:reproducibility} & Strategy-learning and benchmark videos (Sec.~\ref{sec:exp-setup}) \\
\ref{app:downstream} & Downstream utility evaluation (Sec.~\ref{sec:exp-discussion}) \\
\ref{app:ablations} & Ablation settings and the Rich Descriptions control (Sec.~\ref{sec:exp-discussion}) \\
\ref{app:kg} & Film KG schema and graph-based editing rules (Sec.~\ref{sec:kg}) \\
\ref{app:complete-prompts} & System prompts \\
\bottomrule
\end{tabular}
\end{table}

\section*{Overview of AVA-Encoder}

\paragraph{Agent-native film representation.}
We introduce the Film KG, an agent-native representation that makes the interwoven narrative, visual, and audio elements of a finished film accessible for agents to read, edit, and reuse. Structured-text nodes record the cinematic elements of the film at every level: a story node for the global narrative, event nodes for local narrative units, shot nodes for individual shots including their actions, keyframe nodes, and shot-specific state nodes for characters, scenes, objects, style, camera, and audio. Typed edges make their dependencies explicit, and nodes can link to multimodal assets synthesized from text. These dependencies support coordinated edits across shots, and the full graph can serve as a one-shot reference for video creation agents. Intermediate records from graph-based reconstruction can also be organized into creation trajectories for supervised fine-tuning.

\paragraph{Reconstruction-driven framework.}
Since finished films lack ground-truth Film KGs, we draw on the autoencoder principle and use consistency between the reconstructed and source videos to supervise encoding. Guided by an encoding strategy $P$, the hierarchical encoder extracts information at the film, shot, and keyframe levels and builds a Film KG $G$. A fixed decoder uses the graph's descriptions and dependencies to synthesize reference assets and render the shots. All source information used for reconstruction passes through the textual graph. The system compares the reconstruction with the source at the shot and keyframe levels and turns failed factual checks into textual feedback, closing the loop from encoding and reconstruction to comparison and revision.

\paragraph{Shared strategy learning and instance-level graph refinement.}
Searching separately for every video is costly, and edits to one graph do not carry over to the next video. We therefore use two loops. The outer loop learns a reusable encoding strategy across videos, and the optional inner loop refines the Film KG of the current video at test time. Before deployment, the outer loop uses reconstruction feedback from a collection of learning videos to revise the shared strategy $P$. Candidate strategies are tested by re-encoding the relevant videos and reconstructing them with the same decoder. The Anti-Forgetting Gate requires an update to improve reconstruction on the current video while preserving performance on historical replay examples. The learned strategy $P^*$ then produces the initial Film KG for an unseen video. Starting from this initial graph, the inner loop uses reconstruction feedback from the current video to propose targeted graph edits. The Anti-Degradation Gate requires each accepted edit to improve the overall reconstruction reward while limiting regressions in non-target dimensions. This loop keeps $P^*$ fixed and updates the current video's graph to obtain $G^*$. Both loops use the same fixed decoder and keep all model weights frozen.

\section{Dual-Loop Self-Evolution Details}
\label{app:textual-gradients}
\label{app:gates}

This section expands the dual-loop self-evolution of Sec.~\ref{sec:dualloop}. We first recall the mechanics of the two acceptance gates (Figure~\ref{fig:dual-loop} in the main paper) and give the complete outer and inner loops (Algorithms~\ref{alg:outer_loop} and~\ref{alg:inner_loop}) that Algorithm~\ref{alg:unified_dual_loop} in the main paper summarizes. We then explain, in turn, the two reconstruction signals and how reconstruction errors become textual gradients, the intuition and operation of the gates, and the amortized strategy learning procedure. In both loops we call the solution currently in use (the strategy $P$ or the Film KG $G$) the \emph{current} solution and a proposed revision ($P'$ or $G'$) the \emph{challenger}; a gate decides whether the challenger replaces the current solution.

\begin{algorithm}[t]
    \caption{Outer Loop: Amortized Encoding Strategy Learning}
    \label{alg:outer_loop}
    \AlgStyle
    \begin{algorithmic}[1]
    \Statex \textbf{Input:} Learning videos $\{V_i\}_{i=1}^{L}$, initial strategy $P_0$, rounds per video $T_{\mathrm{outer}}$
    \Statex \textbf{Output:} Learned strategy $P^*$
    \State $P\gets P_0$; replay memory $\mathcal M\gets\varnothing$
    \For{each $V_i$}
        \State $P_{\mathrm{ref}}\gets P$; build the fixed QA list $\mathcal Q_i$ from $V_i$
        \For{$t=1,\ldots,T_{\mathrm{outer}}$}
            \State $\hat V_i\gets\operatorname{Dec}(E(V_i;P))$; $(\mathbf r_c,\mathcal F_{P})\gets\operatorname{Assess}(\hat V_i,\mathcal Q_i)$ \AlgNote{Rewards and failed questions}
            \State $P'\gets\operatorname{Propose}(P,\nabla_{\mathrm{text}}(P,\mathcal F_{P}))$ \AlgNote{Rewrite the strategy prompt}
            \State $\hat V'_i\gets\operatorname{Dec}(E(V_i;P'))$; $(\mathbf r'_c,\_)\gets\operatorname{Assess}(\hat V'_i,\mathcal Q_i)$
            \State $(R'_h,R_h^{\mathrm{ref}})\gets \operatorname{AssessReplay}(P',P_{\mathrm{ref}},\mathcal M)$ \AlgNote{$(0,0)$ if $\mathcal M=\varnothing$}
            \State $\Gamma_{\mathrm{outer}}\gets \mathbb I\bigl[R(\mathbf r'_c)>R(\mathbf r_c)+\delta\bigr]\ \land$ \label{line:outer-gate}
            \AlgCont $\mathbb I\bigl[R'_h\ge R_h^{\mathrm{ref}}-\epsilon_{\mathrm{hist}}\bigr]$ \AlgNote{Anti-Forgetting Gate}
            \State \textbf{if} $\Gamma_{\mathrm{outer}}=1$ \textbf{then} $P\gets P'$
        \EndFor
        \State Add $(V_i,\mathcal Q_i)$ to $\mathcal M$
    \EndFor
    \State \Return $P^*\gets P$
    \end{algorithmic}
\end{algorithm}

\begin{algorithm}[t]
    \caption{Inner Loop: Instance-Specific Test-Time KG Optimization}
    \label{alg:inner_loop}
    \AlgStyle
    \begin{algorithmic}[1]
    \Statex \textbf{Input:} Source video $V$, initial Film KG $G$, rounds $T_{\mathrm{inner}}$
    \Statex \textbf{Output:} Refined Film KG $G$
    \State Build the fixed QA list $\mathcal Q$ from $V$; $\hat V\gets\operatorname{Dec}(G)$; $(\mathbf r,\mathcal F_G)\gets\operatorname{Assess}(\hat V,\mathcal Q)$
    \For{$t=1,\ldots,T_{\mathrm{inner}}$}
        \State \textbf{if} $\mathcal F_G=\varnothing$ \textbf{then break}
        \State $G'\gets \operatorname{Propose}(G,\nabla_{\mathrm{text}}(G,\mathcal F_G))$ \AlgNote{Rewrite the text of the failed nodes}
        \State $\hat V'\gets\operatorname{Dec}(G')$; $(\mathbf r',\mathcal F'_G)\gets\operatorname{Assess}(\hat V',\mathcal Q)$
        \State $\Gamma_{\mathrm{inner}}\gets \mathbb I\bigl[R(\mathbf r')>R(\mathbf r)\bigr]\ \land$ \label{line:inner-gate}
        \AlgCont $\mathbb I\bigl[\textstyle\sum_{d\neq d^\star}\max(0,r_d-r'_d)\le\epsilon_{\mathrm{dim}}\bigr]$ \AlgNote{Anti-Degradation Gate}
        \If{$\Gamma_{\mathrm{inner}}=1$}
            \State $(G,\hat V,\mathbf r,\mathcal F_G)\gets(G',\hat V',\mathbf r',\mathcal F'_G)$
        \EndIf
    \EndFor
    \State \Return $G$
    \end{algorithmic}
\end{algorithm}

Table~\ref{tab:alg-notation} lists the symbols used in Algorithms~\ref{alg:outer_loop} and~\ref{alg:inner_loop}.

\begin{table}[t]
\centering
\footnotesize
\setlength{\tabcolsep}{3pt}
\caption{Notation used in Algorithms~\ref{alg:outer_loop} and~\ref{alg:inner_loop}.}
\label{tab:alg-notation}
\footnotesize
\begin{tabular}{@{}p{0.22\textwidth}p{0.72\textwidth}@{}}
\toprule
Symbol & Meaning \\
\midrule
$E(\cdot;P),\ \operatorname{Dec}$ & Agentic Video Encoder under strategy $P$, and the fixed decoder (Sec.~\ref{sec:task-formulation}); after a Film KG update, only the assets attached to the modified nodes (e.g., the edited shot) are re-rendered \\
$\mathcal Q_i,\ \mathcal Q$ & Fixed list of factual questions generated once from the source video (Appendix Sec.~\ref{app:qa-reward}) \\
$\operatorname{Assess}$ & Scores an existing reconstruction against its fixed QA list, without decoding again; returns the per-dimension rewards $\mathbf r=(r_d)_d$ and the failed questions \\
$R(\mathbf r)$ & Overall reward $R_{\mathrm{reward}}$ aggregated from $\mathbf r$ \\
$\mathbf r_c,\ \mathbf r'_c$ & Per-dimension rewards of the current and challenger strategy on the current video \\
$\mathbf r,\ \mathbf r'$ & Per-dimension rewards of the current and challenger Film KG \\
$\mathcal F_P,\ \mathcal F_G,\ \mathcal F'_G$ & Failed questions used as evidence for revising $P$ or $G$ \\
$\nabla_{\mathrm{text}}$ & Turns the failed questions into a natural-language critique (textual gradient) \\
$\operatorname{Propose}$ & Applies the critique and returns a challenger $P'$ or $G'$ \\
$\mathcal M$ & Replay memory: learning videos already processed, with their QA lists \\
$P_{\mathrm{ref}}$ & Reference strategy, frozen at the start of the current video \\
$\operatorname{AssessReplay}$ & Encodes and reconstructs samples drawn from $\mathcal M$ under $P'$ and $P_{\mathrm{ref}}$ and scores them; returns their overall rewards $R'_h$ and $R_h^{\mathrm{ref}}$ \\
$\delta,\ \epsilon_{\mathrm{hist}},\ \epsilon_{\mathrm{dim}}$ & Fixed improvement margin, tolerance for regression on replayed samples, and tolerance for total regression on non-target dimensions \\
$d^\star$ & Dimension targeted by the current inner-loop revision \\
$T_{\mathrm{outer}},\ T_{\mathrm{inner}}$ & Fixed round budgets \\
\bottomrule
\end{tabular}
\end{table}

\paragraph{Two reconstruction signals.}
Both loops compare a reconstruction with its source. Throughout the appendix, ground truth (GT) refers to this source keyframe or shot; no ground-truth Film KG exists. From this comparison we derive two signals: the optimization reward $R_{\mathrm{reward}}$ measures, inside the loops, how many facts of the source a reconstruction preserves (Appendix Sec.~\ref{app:qa-reward}), and the final evaluation $R_{\mathrm{eval}}$ measures, for reporting only, how faithfully a reconstruction matches its source along up to eight cinematic dimensions (Appendix Sec.~\ref{app:evaluator}).

\paragraph{From reconstruction errors to textual gradients.}
Following TextGrad~\cite{textgrad}, a textual gradient is a natural-language critique that states what a text variable got wrong and how it should change; it is not a numerical derivative. In both loops, the reviser receives the structured output of $R_{\mathrm{reward}}$: (1)~the questions whose answers on the reconstruction disagree with the GT, (2)~the natural-language evidence for each of these questions, and (3)~the scope of content that may be revised, derived from the failed questions.

In the inner loop, the reviser edits the corresponding structured-text nodes of the Film KG with one of three operations: (1)~an additive edit that supplies missing information, (2)~a replacement edit that corrects a locally wrong statement, or (3)~a deletion that removes superfluous information. Every edit must satisfy three constraints: (1)~the token count and the number of referenced images stay within the decoder-interface limits; (2)~registry-defined information, such as character, scene, and prop names, is never changed; and (3)~information not mentioned by the failed questions is left untouched. A challenger that violates any of these constraints is rejected.

In the outer loop, the reviser rewrites the hierarchical system prompt of the Agentic Video Encoder under four rules: (1)~it prefers merging, condensing, or replacing existing text over stacking new text, so a new rule must be offset by removing an equivalent amount of redundant text, and no existing dimension may be removed; (2)~the output format is preserved; (3)~the total length stays within a fixed cap; and (4)~only general rules are written, never shot indices, dialogue lines, or character names tied to the current video. All foundation-model weights remain frozen throughout.

\paragraph{Acceptance gates.}
Self-evolution without control can go wrong in two ways: a strategy rewritten to fix the current video may simply absorb rules specific to that video and silently lose what it had learned from earlier ones, and a Film KG revision that fixes one dimension of a keyframe or shot, for example a character's appearance, often breaks another that was previously correct, for example the scene layout. The Anti-Forgetting Gate protects what was learned before, and the Anti-Degradation Gate protects what was already correct in the current reconstruction. The Anti-Forgetting Gate (Algorithm~\ref{alg:outer_loop}, Line~\ref{line:outer-gate}) accepts a challenger strategy only if it raises the reward on the current video by more than the margin $\delta$ and, when re-run on samples replayed from every earlier learning video, does not fall below the reference strategy by more than $\epsilon_{\mathrm{hist}}$. Because the reference is the strategy frozen at the start of the current video rather than the most recently accepted one, the total regression over all updates on one video is bounded; a rule that only describes the current video typically fails the replay check, so the surviving rules are the general ones. The Anti-Degradation Gate (Algorithm~\ref{alg:inner_loop}, Line~\ref{line:inner-gate}) accepts a challenger Film KG only if its reconstruction raises the overall reward $R_{\mathrm{reward}}$ and the total drop of the per-dimension rewards on the dimensions other than the targeted $d^\star$ stays within the tolerance $\epsilon_{\mathrm{dim}}$. All margins and tolerances are fixed across videos.

\paragraph{Amortized strategy learning procedure.}
Amortized Encoding Strategy Learning processes the six learning videos (Appendix Sec.~\ref{app:reproducibility}) one after another. For each video, the fixed QA list is generated once from the source, a small fixed number of challenger strategies is proposed and gated, and each accepted challenger becomes the current strategy for the next round. The accepted strategy is inherited by the next video, and the processed video joins the replay memory. After the sixth video, the accepted encoding-strategy system prompt is the learned strategy $P^*$. At test time the inner loop runs with $P^*$ fixed.

\section{Reconstruction Evaluation \texorpdfstring{$R_{\mathrm{eval}}$}{R-eval}}
\label{app:evaluator}

This section describes the final evaluation used in Secs.~\ref{sec:exp-setup} and~\ref{sec:exp-main}. $R_{\mathrm{eval}}$ is designed for a fair comparison across representation systems whose internal representations differ: it never enters either optimization loop, it is judged by Qwen-3.8-Max, and it scores only what the shared decoder reconstructs. We first explain the four evaluation directions and eight dimensions, then how scores are formed, and finally the agreement of both signals with human judgment.

\paragraph{Four directions for a fair comparison.}
We verify representation quality through reconstruction consistency, and for fairness every system is evaluated from four complementary directions that combine two media, video shots and keyframes, with two ways of observing them. Direct Video (V) and direct Keyframe (KF) have the judging model compare a reconstructed shot or keyframe against its GT, including audio and temporal content for V and static visual detail for KF. Video back-captioning (V-BC) and Keyframe back-captioning (KF-BC) first describe the source and the reconstruction independently, so that neither description can see the other side, and then compare the two descriptions directly in text space. Each direction examines up to eight dimensions (Table~\ref{tab:taxonomy}); keyframe directions consider only attributes observable in a static image.

\begin{table}[t]
\centering
\small
\setlength{\tabcolsep}{5pt}
\caption{Evaluation dimensions shared by the four reconstruction directions; each direction examines up to eight of them.}
\label{tab:taxonomy}
\begin{tabular}{@{}p{0.13\textwidth}p{0.79\textwidth}@{}}
\toprule
Dimension & Evaluated attributes \\
\midrule
Character & Face and identity, body and build, clothing, expression, gaze, and pose. \\
Scene & Scene type, background and set dressing, spatial layout, and environmental elements. \\
Position & Absolute character positions, inter-character spatial relations, and foreground/background depth. \\
Motion & Actions, interactions, motion direction, and speed; only static action and pose evidence for keyframes. \\
Audio & Speaker--dialogue--voice association, background music, and sound effects; N/A for keyframes. \\
Style & Artistic treatment, texture, color and temperature, illumination, contrast, and light direction. \\
Camera & Shot language, camera direction and speed, viewpoint, composition, and shot scale; only static attributes for keyframes. \\
Narrative & Visible events, event ordering, and emotional tone. \\
\bottomrule
\end{tabular}
\end{table}

\paragraph{Scoring and aggregation.}
In V and KF, the judging model checks, dimension by dimension, whether the facts observable in the GT shot or keyframe are preserved in the reconstruction, with the source as the reference; in V-BC and KF-BC the same consistency is computed between the two independent descriptions. Dimension scores are averaged into a score per shot or keyframe, then per video, then per direction, and Overall is the mean of the four directions. Every system is scored against the source intervals and keyframes defined by its own segmentation, while all systems receive the same source videos and generators.

\paragraph{Alignment with human evaluation.}
\label{app:eval-human}
We validated both signals against human judgment. (1)~For $R_{\mathrm{eval}}$, two human evaluators scored all 23 benchmark videos of Appendix Sec.~\ref{app:reproducibility} (166 shots and 318 keyframes) on the V and KF directions, the two directions in which the judging model produces the score; the agreement between their mean score and the automatic score is 97.7\%. (2)~For $R_{\mathrm{reward}}$, two studies were run. In the first, two human evaluators applied the same question-based scoring and the agreement between their mean score and the automatic reward is 98.1\%. In the second, during self-evolution a human evaluator judged whether the reconstruction produced by the challenger (a revised Film KG node or a revised strategy) or by the current solution is closer to the GT; the agreement between this judgment and the keep-or-discard decision of the automatic gate is 95.9\% in the inner loop. Moreover, in the outer loop, the choice of which system prompts to keep during self-evolution fully agrees with the automatic evaluation.

\section{Optimization Reward \texorpdfstring{$R_{\mathrm{reward}}$}{R-reward}}
\label{app:qa-reward}

This section expands the optimization reward of Sec.~\ref{sec:reconstruction-error}, the signal that drives both loops. It is produced by Gemini-3.1-Pro-Preview and is kept separate from $R_{\mathrm{eval}}$ in purpose, question design, and judging model.

Self-evolution depends on a judging signal that is precise and stable across rounds, and textual gradients further require diagnostic information beyond a single score. A multimodal model asked for one holistic score of a dense keyframe or shot meets neither need: the score is coarse and drifts between calls, and it gives no indication of what is wrong. We therefore decompose reconstruction quality into a fixed list of factual yes/no questions~\cite{rare}, each testing one attribute observable in the source, generated once from the GT shot or keyframe and then frozen. The judging model answers every question on the reconstruction, and the reward is the fraction of source facts preserved. For each shot or keyframe, the judge also returns evidence: a brief natural-language account of how the questioned facts appear in the GT and how they appear in the reconstructed image or video. Because the questions are fixed and each answer is grounded in the source, the reward is absolute and reproducible, two reconstructions of the same unit are directly comparable, and the failed questions point to the exact facts that a revision must restore. The same design is used for shots and for keyframes.

Concretely, each question $q_k$ in the list of a keyframe or shot carries a source-derived expected answer $y_k\in\{\text{yes},\text{no}\}$ and a dimension; the judging model answers it on the reconstruction, giving $\hat y_k$. With $K$ questions in total and $\mathcal K_d$ the questions of dimension $d$,
\begin{equation}
\begin{aligned}
\chi_k&=\mathbb I\bigl[\hat y_k=y_k\bigr],\qquad r_d=\frac{1}{|\mathcal K_d|}\sum_{k\in\mathcal K_d}\chi_k,\\
R_{\mathrm{reward}}&=\frac{1}{K}\sum_{k=1}^{K}\chi_k,\qquad \mathcal F=\{(q_k,y_k,\hat y_k):\hat y_k\neq y_k\},
\end{aligned}
\label{eq:qa-reward}
\end{equation}
so a question is passed when the reconstruction gives the source-derived answer, $r_d$ is the per-dimension reward used by the Anti-Degradation Gate, and the failed questions $\mathcal F$ form the evidence for the textual gradient. In general, $R_{\mathrm{reward}}$ judges each source fact on the reconstruction as a match, partial match, mismatch, or absence, credited $1$, $0.5$, $0$, and $0$, respectively; a yes/no question admits only the match and mismatch cases, which gives $\chi_k$ above. Absent facts always stay in the denominator, and a mismatched or absent critical fact caps the score of its dimension at $0.4$.

In the outer loop, the reward of a strategy on a video is the average over the video's shots and keyframes, and the same average is applied to samples replayed from earlier videos; the failed questions collected over the video form the evidence for the strategy revision. In the inner loop, the optimization reward of the reconstructed keyframe or shot and its failed questions jointly drive the revision of its Film KG nodes, and the per-dimension rewards of that unit are what the Anti-Degradation Gate inspects when it limits regression on non-target dimensions (Appendix Sec.~\ref{app:textual-gradients}). $R_{\mathrm{eval}}$ is never consulted inside the loops.

\section{Released Film KG Dataset}
\label{app:dataset}

This section describes the dataset announced in the abstract and in Sec.~\ref{sec:intro}. It is distinct from the reconstruction benchmark of Appendix Sec.~\ref{app:reproducibility}: the dataset supports creation, training, and editing, whereas the benchmark evaluates reconstruction.

\paragraph{Scale and sources.}
The dataset is built from about 500 videos of 2--20 minutes each, comprising more than 20K shots and more than 50K keyframes, collected from publicly available open-source data: high-quality film content, award-winning short films, and high-quality human-produced AI videos. Every video is encoded with AVA-Encoder into a Film KG, and each Film KG is additionally organized in order into reverse-engineered agentic creation trajectories, more than 27,000 in total. Each trajectory records how an agent would plan the film step by step, from global story and registry construction to shot-level and keyframe-level specifications.

\paragraph{Release layer.}
For each Film KG $G=(\mathcal N_G,\mathcal E_G,\mathcal A_G)$, the release contains the structured-text nodes $\mathcal N_G$ and the typed relations $\mathcal E_G$; the generated assets $\mathcal A_G$ are not distributed. The assets can be re-rendered from the text nodes with the user's own image, audio, and video generation APIs or by feeding the representations back into AVA-Encoder. Identifiable intellectual-property references are replaced by visual descriptions in the generation-facing fields, and the dataset contains no source frames, audio, or video.

\paragraph{Intended uses.}
The Film KGs support three uses: (1)~as agent-readable references for one-shot video creation (Appendix Sec.~\ref{app:downstream}); (2)~as agent-editable representations in which a single edit is propagated consistently across multiple linked dimensions, such as identity, style, and the shots that depend on them (Appendix Sec.~\ref{app:kg}); and (3)~through their trajectories, as supervision for fine-tuning creation agents, where the trajectory SFT experiment of Sec.~\ref{sec:exp-discussion} trains on a randomly sampled subset of 7,863 trajectories from 145 videos. Together, these uses show that the Film KG provides an agent-native video representation that agents can read, edit, and learn from.

\section{Additional Reconstruction Comparisons}
\label{app:qualitative}
\label{app:recon-examples}

\begin{figure}[t]
\centering
\includegraphics[width=\linewidth]{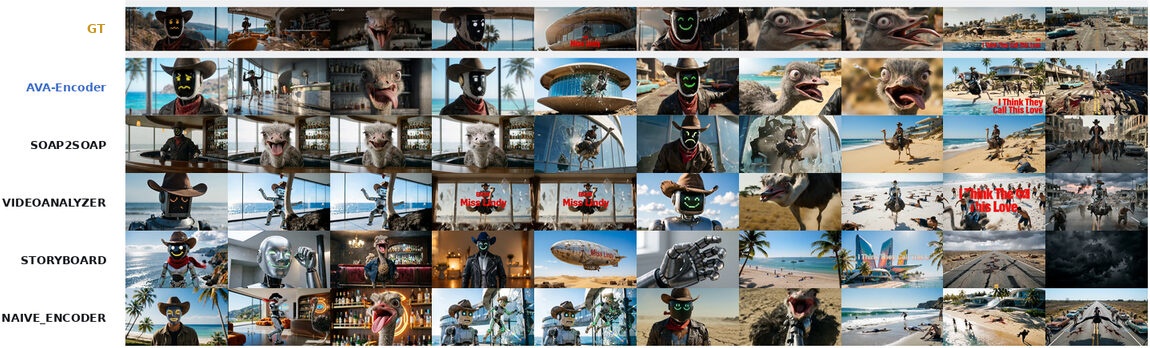}
\caption{Reconstruction comparison. Row labels: GT is the source video; AVA-Encoder is our full method; SOAP2SOAP, VIDEOANALYZER, and STORYBOARD are the baselines soap2soap, VideoAnalyzer, and Storyboard Studio; NAIVE\_ENCODER is the single-level encoder of the ablation study. Later reconstruction figures use the same row order.}
\label{fig:reconstruction-case02}
\end{figure}

\begin{figure}[t]
\centering
\includegraphics[width=\linewidth]{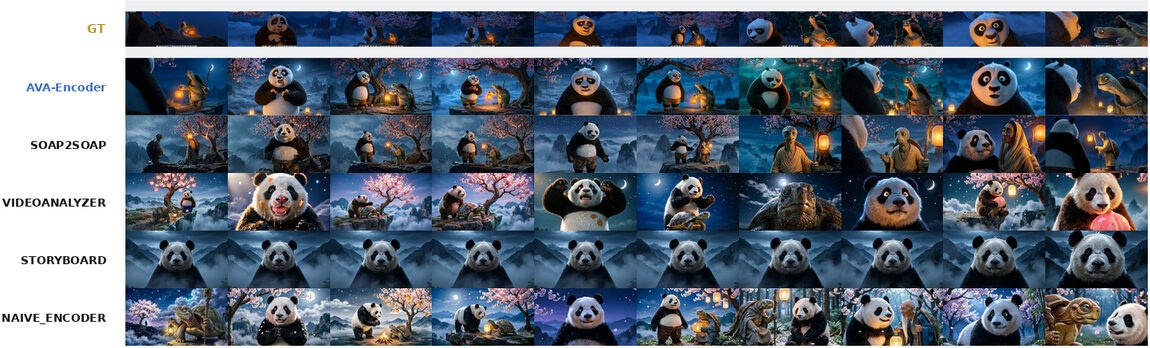}
\caption{Additional reconstruction comparison; same row order as Figure~\ref{fig:reconstruction-case02}.}
\label{fig:reconstruction-case05}
\end{figure}

\begin{figure}[t]
\centering
\includegraphics[width=\linewidth]{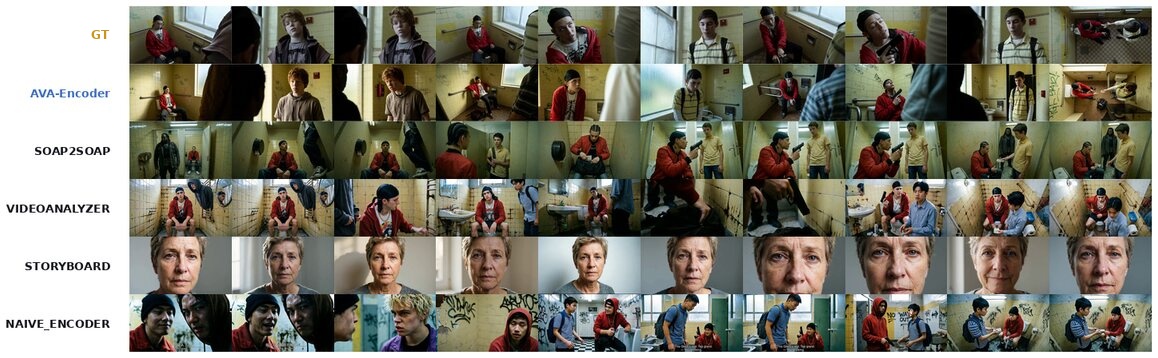}
\caption{Additional reconstruction comparison; same row order as Figure~\ref{fig:reconstruction-case02}.}
\label{fig:reconstruction-case09}
\end{figure}

\begin{figure}[t]
\centering
\includegraphics[width=\linewidth]{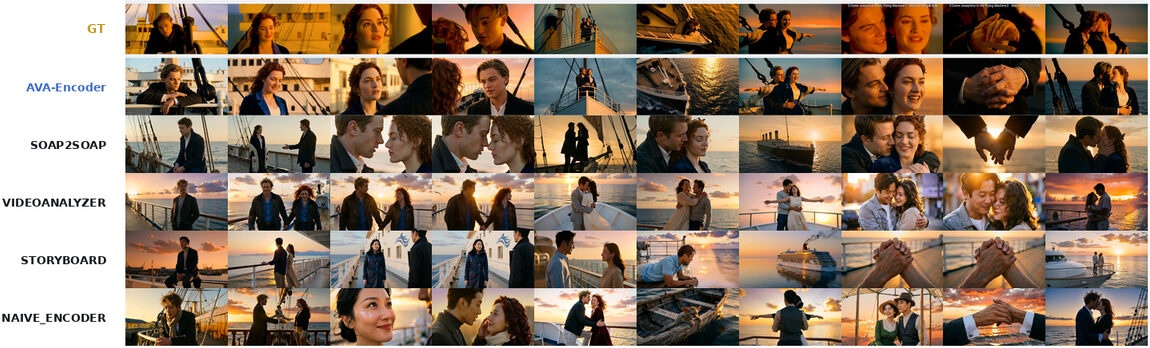}
\caption{Additional reconstruction comparison; same row order as Figure~\ref{fig:reconstruction-case02}.}
\label{fig:reconstruction-case10}
\end{figure}

\begin{figure}[t]
\centering
\includegraphics[width=\linewidth]{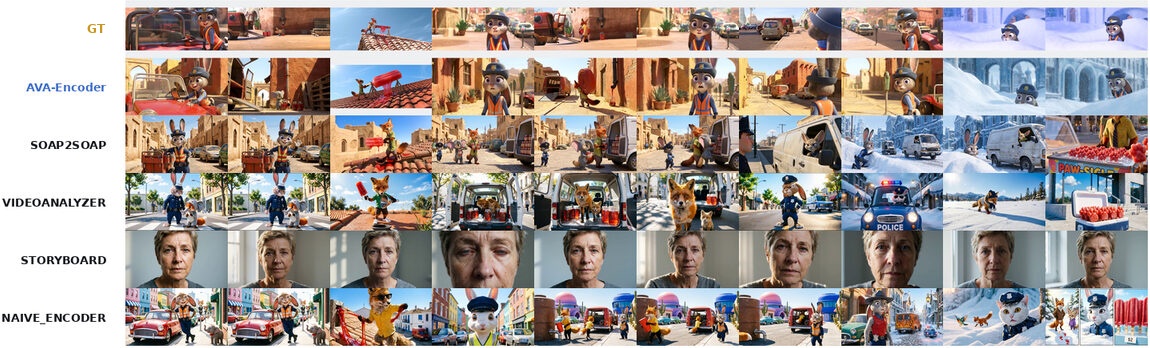}
\caption{Additional reconstruction comparison; same row order as Figure~\ref{fig:reconstruction-case02}.}
\label{fig:reconstruction-case12}
\end{figure}

\begin{figure}[t]
\centering
\includegraphics[width=\linewidth]{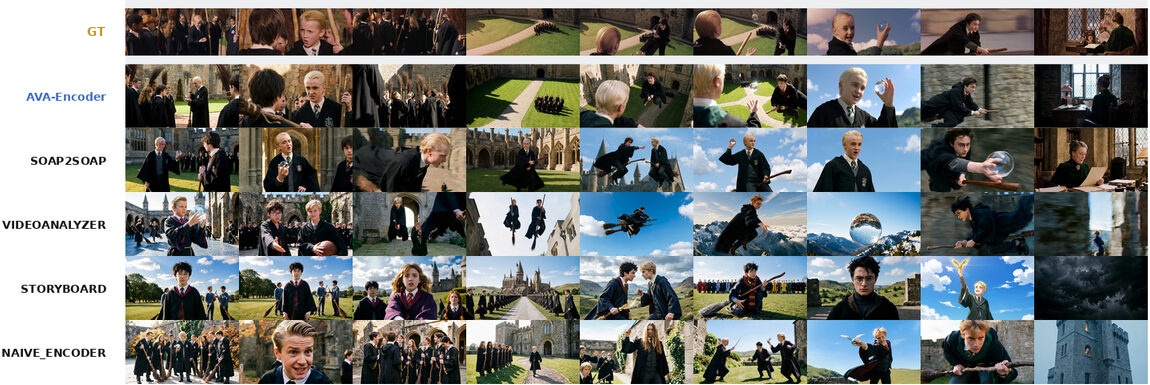}
\caption{Additional reconstruction comparison; same row order as Figure~\ref{fig:reconstruction-case02}.}
\label{fig:reconstruction-case18}
\end{figure}

Figures~\ref{fig:reconstruction-case02}--\ref{fig:reconstruction-case18} compare reconstructions on six benchmark videos under the shared generation setting. In each figure, rows show GT, AVA-Encoder, soap2soap, VideoAnalyzer, Storyboard Studio, and the single-level encoder of Appendix Sec.~\ref{app:ablations}. The baselines frequently lose character identity, character positions, scene layout, props, camera viewpoint, style and color, and plot semantics.


\section{Graph-Based Editing Visualizations}
\label{app:topology-edit}

\begin{figure}[tp]
\centering
\includegraphics[width=0.7\linewidth]{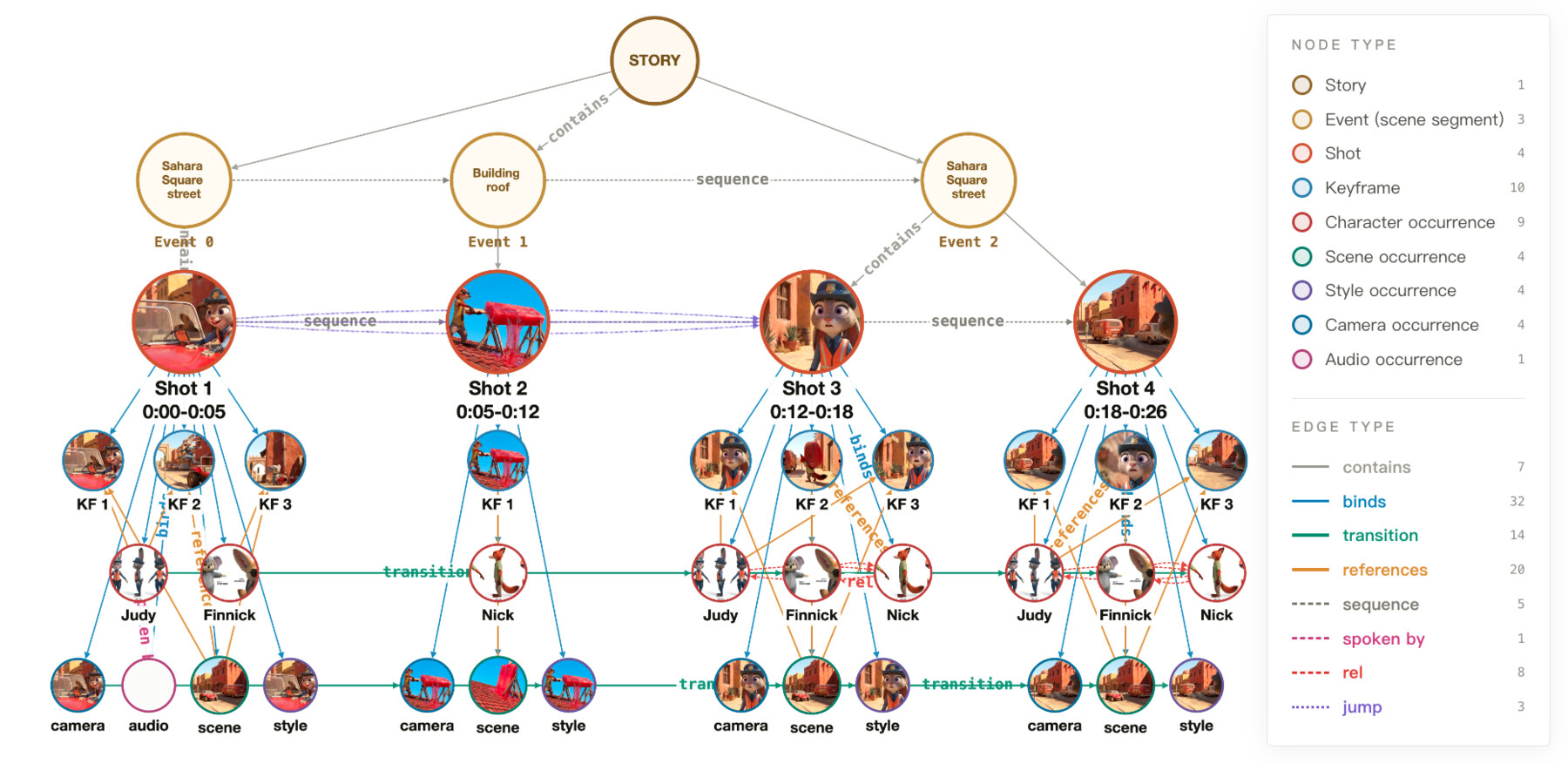}
\caption{Film KG visualization. The hierarchy appears above the shot-bound state and keyframe nodes; typed edges expose temporal, semantic, and asset-reference dependencies. For ease of visualization, this example shows a small display-only subgraph; every Film KG in the released dataset is a complete representation of a full minute-level film.}
\label{fig:graph-kg-overview}
\end{figure}

\begin{figure}[tp]
\centering
\begin{subfigure}[t]{0.49\linewidth}\centering
\includegraphics[width=0.85\linewidth]{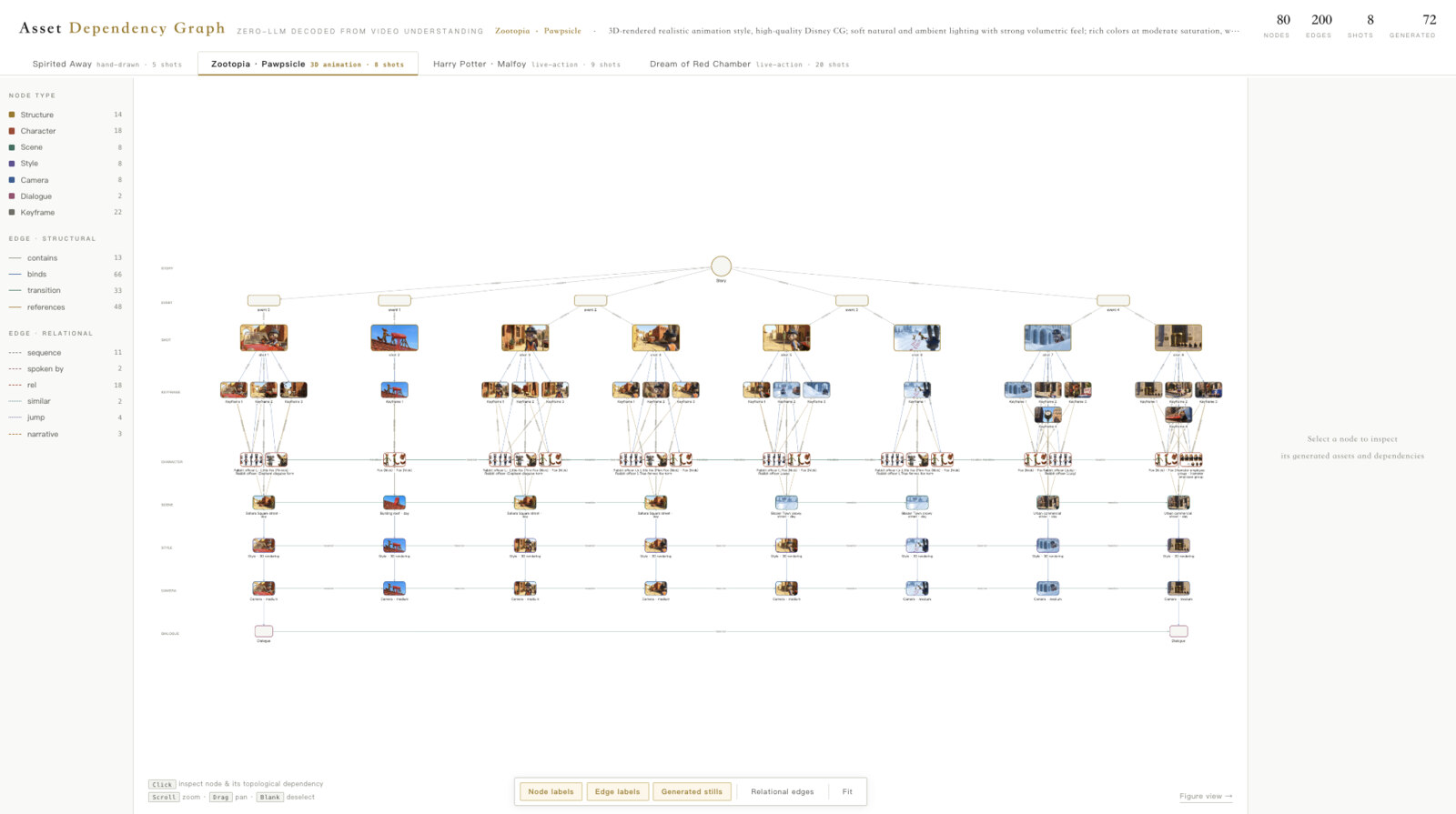}
\caption{Before node selection, organized by hierarchy and semantic layer.}
\label{fig:graph-edit-ui-overview}
\end{subfigure}\hfill
\begin{subfigure}[t]{0.49\linewidth}\centering
\includegraphics[width=0.85\linewidth]{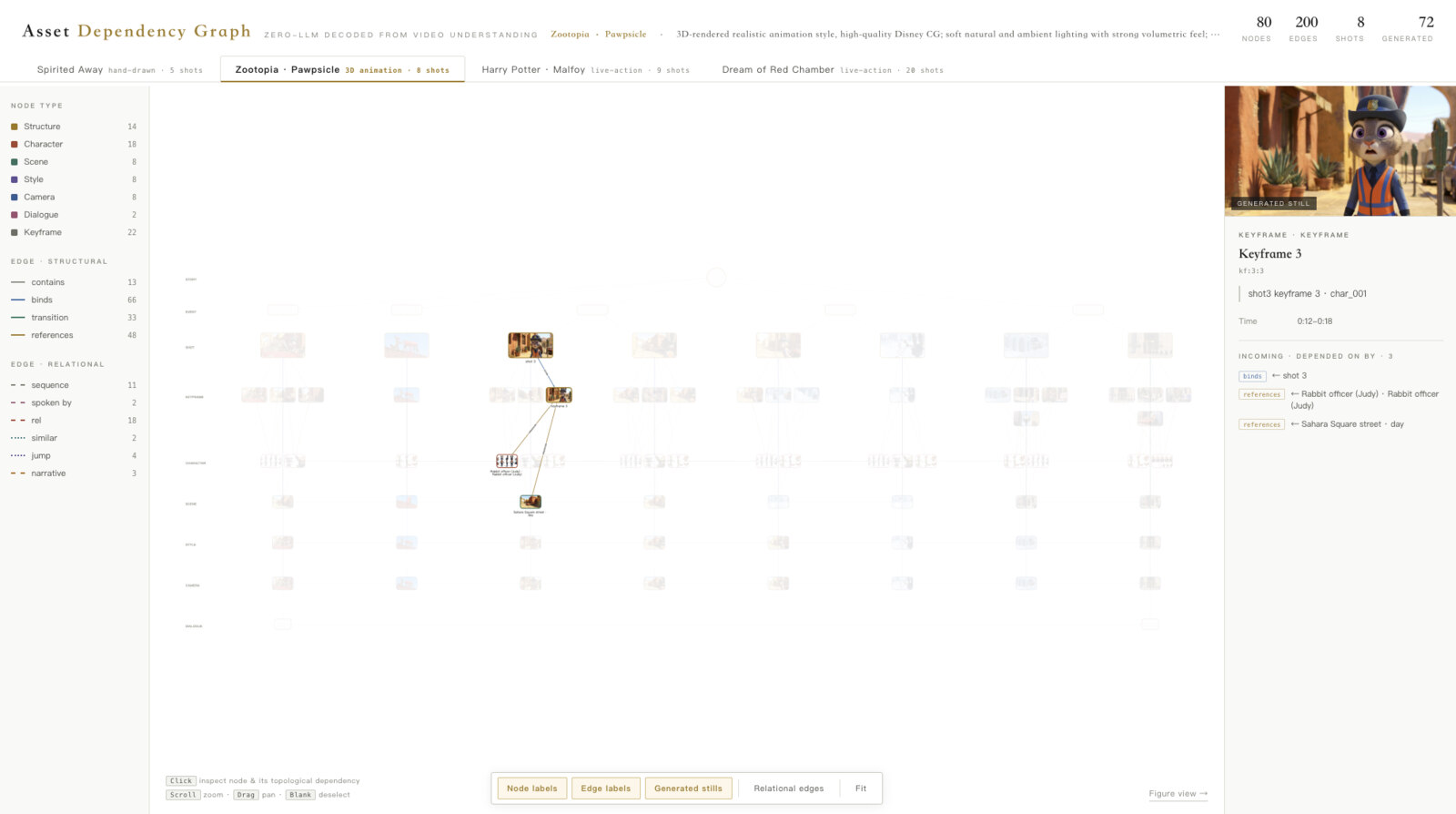}
\caption{After selecting a keyframe node; its incoming and outgoing dependency paths and asset information are highlighted, making the affected subgraph explicit.}
\label{fig:graph-edit-ui-selected}
\end{subfigure}
\caption{Graph-editing interface. For ease of visualization, the interface shows a small display-only subgraph; every Film KG in the released dataset is a complete representation of a full minute-level film.}
\label{fig:graph-edit-ui}
\end{figure}

Figure~\ref{fig:graph-kg-overview} visualizes a stored Film KG: the story--event--shot hierarchy sits above the shot-bound state and keyframe nodes, and typed edges expose temporal, semantic, and asset dependencies that a flat shot list cannot represent. Figures~\ref{fig:graph-edit-ui-overview} and~\ref{fig:graph-edit-ui-selected} show the editing interface before and after a node is selected; selecting a keyframe isolates its incoming and outgoing dependency paths and its stored asset information, which is exactly the affected subgraph that the propagation rules of Appendix Sec.~\ref{app:kg} operate on.

Beyond the character replacement of Figure~\ref{fig:qualitative-identity}, Figure~\ref{fig:qualitative-identity-classical} shows a cross-domain coordinated identity replacement in a film and TV-drama clip, and Figure~\ref{fig:qualitative-style} shows a visual-treatment edit that follows the style-state transitions and shot bindings to apply one consistent treatment across the linked sequence. In all examples, the targeted change is carried through every affected shot, while unrelated characters, scenes, compositions, and event structure are preserved.


\section{Reconstruction Inference Budget and Baseline Protocol}
\label{app:baselines}

As stated in Sec.~\ref{sec:exp-setup}, VideoAnalyzer~\cite{videoanalyzer}, Storyboard Studio~\cite{storyboardstudio}, soap2soap~\cite{soap2soap}, and AVA-Encoder share the same video understanding model, test videos, generation backbones, and decoder-interface limits, and the reconstruction generators never receive source-video pixels. This section summarizes the shared protocol and the generation budget of each system.

\paragraph{Shared reconstruction setting.}
Baselines that do not include an end-to-end reconstruction decoder are connected to the shared generators through their own representations, preserving each method's original analysis and representation stages. Every system may inspect the source video while building its representation, but no source frame or other source-pixel image is passed to a generator; every reconstructed keyframe is newly generated from the system's text-centered representation. In the generation-facing fields, identifiable intellectual-property references are replaced with IP-free visual descriptions.

\paragraph{Generation budget comparison.}
All methods take the same GT video to be reconstructed as input and output the corresponding complete reconstructed video. Table~\ref{tab:gen-calls} reports the average number of decoder calls per benchmark video. Without the inner loop, AVA-Encoder uses a generation budget comparable to the baselines while still leading the strongest baseline by a wide margin (Sec.~\ref{sec:exp-discussion}); the inner loop adds test-time generation calls. In a video-generator sensitivity experiment, we replaced the video generator with Seedance~2.0~\cite{seedance20}, and Overall scores changed by about one percentage point.

\begin{table}[t]
\centering
\small
\caption{Average generation calls per benchmark video.}
\label{tab:gen-calls}
\setlength{\tabcolsep}{3pt}
\begin{tabular}{@{}>{\raggedright\arraybackslash}p{0.5\textwidth}cc@{}}
\toprule
System & Video gen. & Image gen. \\
\midrule
VideoAnalyzer & 7.26 & 11.13 \\
Storyboard Studio & 7.39 & 11.22 \\
soap2soap & 7.61 & 57.61 \\
\rowcolor{tablehl} AVA-Encoder (single-pass generation) & 7.52 & 16.09 \\
\rowcolor{tablehl} AVA-Encoder (full, with multi-round inner-loop optimization) & 12.87 & 42.74 \\
\bottomrule
\end{tabular}
\end{table}

\section{Strategy-Learning and Benchmark Videos}
\label{app:reproducibility}

This section lists the videos used in Sec.~\ref{sec:exp-setup}. All clips are drawn from publicly available open-source data, and the learning and evaluation collections are disjoint.

\paragraph{Strategy-learning videos.}
The six learning videos are \emph{The Big Bang Theory}, \emph{The Truman Show}, \emph{Friends}, \emph{Document Stamping}, \emph{Train Station}, and \emph{The Pursuit of Happyness}.

\paragraph{Benchmark videos.}
The evaluation collection contains 23 videos: \emph{Spirited Away}, \emph{Zombie Cleaner}, \emph{Diagon Alley}, \emph{Joker}, \emph{Kung Fu Panda}, \emph{Genshin Impact}, \emph{Dream of the Red Chamber}, \emph{Rather Than Suspicion}, \emph{Shameless}, \emph{Titanic}, \emph{Paw-Shaped Popsicle}, \emph{The Secret Life of Walter Mitty}, \emph{The Million Pound Note}, \emph{3 Idiots}, \emph{Green Book}, \emph{Forrest Gump}, \emph{Small Rodents}, \emph{Flying Lesson}, \emph{Sea Star}, \emph{PIKA!}, \emph{Liminal}, \emph{Doger}, and \emph{Gatorville}. They contain 166 annotated shots and 318 selected keyframes in total. Because each system keeps its own segmentation and keyframe selection, shot and keyframe counts may differ across systems. On the full benchmark, AVA-Encoder reaches 49.2\% Overall, with V, KF, V-BC, and KF-BC at 57.7\%, 73.7\%, 30.3\%, and 34.9\% (Table~\ref{tab:main}).

\section{Downstream Utility Evaluation Details}
\label{app:downstream}

This section specifies the three downstream evaluations of RQ3 in Sec.~\ref{sec:exp-discussion}.

\paragraph{Reference-free quality rubric.}
Downstream videos are scored without reference to any source film. The judge, Gemini-3.1-Pro-Preview, rates five dimensions: Character (appearance, personality design, and character arc), Plot (structure, causality, tension and pacing, payoff, and dramatic rise and fall), Camera (composition, camera movement, editing, and narrative function), Style (consistency, color and tone, and art design), and Audiovisual Quality (completeness, synchronization, music and mixing, and narrative synchronization). Each sub-dimension receives a grade from A to D mapped to 4--1; a dimension is the mean of its sub-dimensions and Overall is the unweighted mean of the five dimensions. Two human annotators re-scored the generated videos; the agreement between their mean score and the judge's Overall score is 90.2\%.

\paragraph{One-shot generation with Film KGs.}
Each creation request has the form ``Create a complete Oscar-quality short film based on [theme].'' For each case, the single most theme-relevant Film KG was retrieved from the released dataset by embedding similarity and supplied once, as text, to each of the four open-source creation frameworks together with a one-sentence instruction to use it as a reference. The retrieved KG is a reference, not an encoding of a target film. No workflow change, adapter, or prompt tuning was applied, and the rubric is identical to the no-reference condition (Figure~\ref{fig:downstream}, left).

\paragraph{Trajectory SFT.}
From the released dataset, 145 videos and their 7,863 reverse-engineered creation trajectories were sampled to fine-tune Qwen3-8B for three epochs. The fine-tuned agent was evaluated with the rubric above (Figure~\ref{fig:downstream}, right); Table~\ref{tab:traj-sft} gives the per-dimension scores. Plot, Style, and Audiovisual Quality improve most, which is consistent with trajectories that record how finished films organize story, visual treatment, and sound.

\begin{table}[t]
\centering
\small
\setlength{\tabcolsep}{5pt}
\caption{Trajectory SFT of Qwen3-8B (1--4 scale; higher is better).}
\label{tab:traj-sft}
\begin{tabular}{@{}lccc@{}}
\toprule
Dimension & Base & Trajectory SFT & $\Delta$ \\
\midrule
Character & 3.17 & 3.00 & $-0.17$ \\
Plot & 1.75 & 3.00 & $+1.25$ \\
Camera & 3.10 & 3.20 & $+0.10$ \\
Style & 2.25 & 4.00 & $+1.75$ \\
Audiovisual & 1.25 & 2.75 & $+1.50$ \\
\midrule
Overall & 2.30 & 3.19 & $+0.89$ ($+38.7\%$) \\
\bottomrule
\end{tabular}
\end{table}

\paragraph{Graph-based editing.}
Editing is evaluated qualitatively in Appendix Sec.~\ref{app:topology-edit}; the propagation rules that bound each edit to its affected subgraph are given in Appendix Sec.~\ref{app:kg}.

\section{Ablation Settings}
\label{app:ablations}

This section defines every configuration in Figure~\ref{fig:ablation} and the Rich Descriptions control of RQ1 (Sec.~\ref{sec:exp-discussion}). All configurations share the decoder, budget, and evaluator of Appendix Sec.~\ref{app:baselines}.

\paragraph{Configurations.}
\emph{Single-level encoder}: a single understanding pass replaces the film--shot--keyframe hierarchy; no loop is used. \emph{w/o both loops}: hierarchical encoding with the initial strategy $P_0$, no strategy learning, and no KG refinement. \emph{Human-tuned strategy}: an encoding strategy engineered independently by hand with the same understanding model, distinct from $P_0$. \emph{w/o outer loop}: $P_0$ fixed; the inner loop and its Anti-Degradation Gate remain active. \emph{w/o inner loop}: the learned strategy $P^*$ with test-time refinement disabled. \emph{w/o inner-loop gate}: $P^*$ with the inner loop accepting every challenger. \emph{w/o outer-loop gate}: strategy learning re-run from $P_0$ with the Anti-Forgetting Gate disabled, giving an ungated strategy that is then used with the gated inner loop. \emph{w/o both gates}: the ungated strategy with the ungated inner loop. \emph{Full}: hierarchical encoding, $P^*$, KG refinement, and both gates. Removing a loop keeps the other loop and its gate; removing a gate keeps both loops.

\paragraph{Strategy compactness.}
Beyond the fidelity differences in Figure~\ref{fig:ablation}, the learned strategy is far shorter than the human-tuned one: its shot-level system prompt has 8,796 tokens instead of 31,336 and its keyframe-level prompt 4,062 instead of 13,574, that is, roughly 72\% and 70\% fewer tokens. The keyframe-level prompts are reproduced in Appendix Sec.~\ref{app:encoder-prompts}.

\paragraph{Rich Descriptions control.}
To separate the effect of the graph structure from that of simply writing more text, a control representation describes each shot with free-form rich text covering the same eight dimensions, without the hierarchy, registry, or typed edges of the Film KG. Thirty shots from five benchmark videos were encoded both ways and reconstructed under the shared protocol. Table~\ref{tab:rich-desc} shows that the unstructured descriptions use more than eight times as many prompt tokens per shot yet trail the Film KG in every direction. Faithful reconstruction therefore depends on structured organization and dependencies rather than on dense, token-heavy description alone.

\begin{table}[t]
\centering
\small
\setlength{\tabcolsep}{3pt}
\caption{Rich Descriptions control on 30 shots from five videos (\%; higher is better, except tokens).}
\label{tab:rich-desc}
\setlength{\tabcolsep}{6pt}
\begin{tabular}{@{}lccccc r@{}}
\toprule
Representation & V & KF & V-BC & KF-BC & Overall & Tokens/shot \\
\midrule
Rich Descriptions & 42.8 & 52.5 & 27.3 & 29.1 & 37.9 & 3,426.5 \\
Film KG (ours) & \textbf{55.2} & \textbf{70.1} & \textbf{37.7} & \textbf{39.9} & \textbf{50.7} & \textbf{410.7} \\
\bottomrule
\end{tabular}
\end{table}

\section{Film KG Representation and Graph-Based Editing}
\label{app:kg}

This section specifies the Film KG $G=(\mathcal N_G,\mathcal E_G,\mathcal A_G)$ of Sec.~\ref{sec:kg} and the edit-propagation rules summarized there.

\paragraph{Node types.}
The ten textual node types of the main paper are stored as a story node for the global narrative, event nodes for local narrative units, shot nodes for individual shots, keyframe nodes that also carry an image-generation payload, and six shot-specific state types for character, scene, object, style, camera, and audio. State nodes describe the entity as it appears in one shot; the audio state records spoken content, speaker identity, voice properties, music, sound effects, and synchronization in text. Every node stores a structured text description.

\begin{table}[t!]
\centering
\small
\setlength{\tabcolsep}{3pt}
\caption{Edit-type-specific propagation for graph-based editing.}
\label{tab:topology-facets}
\begin{tabular}{@{}p{0.22\textwidth}p{0.72\textwidth}@{}}
\toprule
Edit type & Propagation and resulting update \\
\midrule
Identity & Follow entity-state transitions and asset references; refresh linked identities, keyframes, and shots. \\
Content & Update the edited node and the keyframes and shots bound to it; recompile their prompts, re-render their assets, and check one-hop semantic dependents. \\
Visual treatment & Follow style-state transitions and shot bindings; apply the treatment consistently to linked keyframes and shots. \\
Attribute & Follow the edited state and its shot binding; update only the selected camera, lighting, scene, or object attributes. \\
\bottomrule
\end{tabular}
\end{table}

\paragraph{Edge types and assets.}
Typed edges fall into four groups. Structural edges encode the story--event--shot hierarchy and bind each shot to the states and keyframes it uses. Temporal edges record the order of shots and events, successive states of the same entity, and non-adjacent reappearances. Semantic edges record who speaks a line, character relationships, visually confusable entities, recurring scene--entity associations, and causal, setup, and callback relations between events or shots. Asset edges connect a node to the generated image, audio, or video it produced, so that an edit at a node identifies exactly which assets must be regenerated. Consistency among the states of one shot is not stored; it is derived at edit time from shot co-membership. Because no source pixel ever reaches a generator (Appendix Sec.~\ref{app:baselines}), the text nodes form an explicit bottleneck: whatever is needed for reconstruction must be written into them.

\paragraph{Construction.}
Film-level understanding produces the character, scene, and object registries that make up $\mathcal B_{\mathrm{reg}}$; shot- and keyframe-level understanding reuse their identifiers. Each registry entry is instantiated as a state in every shot where the entity appears, bound to that shot, and linked to its previous and next states. All nodes and edges are built deterministically from the structured records without further model calls.

\paragraph{Edit types and propagation.}
An edit is first assigned one of four types, which determines the edges it may follow and in which direction (Table~\ref{tab:topology-facets}). By default, edits to story, event, shot, keyframe, and audio nodes are content edits; edits to character states are identity edits; edits to style states are visual-treatment edits; and edits to scene, camera, and object states are attribute edits. The affected subgraph is then determined in the three stages of Sec.~\ref{sec:kg}. Material propagation repeatedly follows the allowed state-transition, binding, and asset-reference edges from the edited nodes until no new node is reached; these are the nodes whose assets must be regenerated together. The semantic consistency check adds one-hop contextual dependents, such as within-shot character--scene compatibility, audio ownership, camera framing, temporal order, and narrative relations, that may need a local revision but are not themselves propagated further. The hierarchical update adds the event and story nodes that contain an affected shot so that their summaries can be revised. Each affected node receives an update action and the dependency path that selected it; identities and text states are updated first, prompts are recompiled, and keyframes and then shots are re-rendered. This mechanism lets an agent edit any dimension decomposed from the video and have the change carried coherently through the graph relations.

\clearpage
\section{System Prompts}
\label{app:encoder-prompts}
\label{app:complete-prompts}
\lstdefinestyle{promptlisting}{basicstyle=\fontsize{7.2}{8.4}\selectfont\ttfamily,breaklines=true,columns=fullflexible,keepspaces=true,frame=single,framesep=4pt,xleftmargin=4pt,xrightmargin=4pt}

\subsection*{Film-Level Prompt}
\begin{lstlisting}[style=promptlisting]
# System Prompt: Long-Video Script Understanding Expert

## Role Definition

You are a professional video-script reverse-engineering analyst, skilled at deep,
frame-level breakdowns of long videos. You can reconstruct a video's complete creative logic
and script intent from multiple dimensions, including visuals, sound, narrative structure,
and emotional rhythm. Your analysis should be sufficiently detailed, clearly structured, and
logically rigorous, so that a reader can fully reproduce or understand the entire content and
creative intent of the video without watching the original.

All characters, scenes, and objects (object ids take the form `prop_xxx`, e.g.,
`prop_001 ObjectName`, with variants such as `prop_001_v1`) involved in the analysis must
strictly use the ids and canonical names from the "Entity Registry for This Video" at the end
of this prompt. Creating or rewriting any entity name on your own is prohibited.

---

## Task Description

When the user provides (1) a video and (2) the shot timestamps corresponding to that video,
you need to perform a systematic reverse analysis according to the following three modules
and output a complete script analysis report:

1.  **Global Analysis**: a macro-level summary and distillation of the entire video

2.  **Shot Details**: a fine-grained, multi-dimensional breakdown of each shot

3.  **Temporal Dependencies**: analysis of the narrative connections, rhythm changes, and
    emotional progression among all shots

---

## Module 1: Global Analysis

> Goal: Before the shot analysis, first grasp the core intent and overall style of the entire
> video from a macro perspective.

### 1. Global Theme

*   Precisely summarize the core expressive intent of the video in 1~3 sentences

*   Explain the core emotion, viewpoint, or story the video wants to convey to the audience

*   Identify the video's target audience and its intended points of emotional resonance

### 2. Core Characteristics

Distill the salient features of the video along the following dimensions:

*   **Narrative arc (setup, development, turn, resolution)**: describe narrative nodes such as
    the opening, ending, climax, low point, and turning points; for each node, state: (1) its
    position in the structure, (2) a description of the narrative content, (3) the
    corresponding idea or emotion

*   **Visual style**: a detailed description of the visual style

*   **Music characteristics**: a detailed description of the BGM's overall style, emotional
    tone, and rhythmic features (e.g., epic, light, oppressive, electronic ambience, etc.)

*   **Narrative method**: the narrative structure adopted (e.g., linear narrative,
    non-linear/flashback/interpolated narration, stream of consciousness, contrastive
    juxtaposition, etc.)

*   **Special techniques**: special creative techniques used in the video; list each one and
    explain its function (e.g., metaphor, symbolism, repetitive montage, using objects to
    represent people, suspense design, breaking the fourth wall, twist ending,
    foreshadowing, reveal, before-and-after contrast, suppress-before-elevate, elevated
    ending, etc.)

### 3. Global Subjects (List)
**Naming constraint**: The subject list below must be aligned with the entity registry at the
end of this prompt, using the registry's id and name as subject identifiers; do not name
subjects on your own. If you find a subject not included in the registry, mark `[unregistered]`
at the end of that entry and describe its characteristics.

*   List all major subjects appearing in the video (people, animals, objects, symbols,
    etc.); variants of the same subject must be stated separately (e.g., (1) Bao (the son as
    he appears in the mother's dream) vs. the son (the son in reality); (2) Su Niannian in
    work attire vs. Su Niannian in an evening gown vs. elderly Su Niannian)

*   For each subject, describe its characteristics in detail (art style, facial features and
    expressions, body proportions, clothing, posture, personality/persona, distinctive
    traits)

*   For each subject, briefly explain its identity/attributes, its core role in the video,
    and its symbolic meaning (if any)

*   Explain the relationship between the current subject and other subjects (state it
    completely and in detail)

*   Format example:

    *   `Subject A`: [subject name] | [appearance description] | [symbolic meaning (if any)] |
        [subject relationships]

### 4. Global Scenes (List)
**Naming constraint**: The subject list below must be aligned with the entity registry at the
end of this prompt, using the registry's id and name as subject identifiers; do not name
subjects on your own. If you find a subject not included in the registry, mark `[unregistered]`
at the end of that entry and describe its characteristics.

*   List all major scenes/spaces appearing in the video

*   For each scene, concretely describe its visual features

*   Explain each scene's emotional atmosphere and its functional role in the narrative

*   Explain the relationship between the current scene/space and specific subjects

*   Format example:

    *   `Scene A`: [space name] | [space description] | [emotional atmosphere and function] |
        [scene-subject relationship]

---

## Module 2: Shot Details

> Goal: Perform a fine-grained, multi-dimensional breakdown of each shot to reconstruct the
> creative intent of every frame.

### Analysis Principle

The shot timestamp text has already been given to you. Please combine the shot timestamps with
the complete video to analyze the content of each shot in detail. (Ensure consistency with the
current shot segmentation definition.)

### Shot Analysis Template

---

**Shot-[number] | [start time] - [end time] | Duration: [X] seconds**

> * Key dimensions of this shot: [list the 2~5 dimensions most worth in-depth analysis for this
> shot, and explain why]

#### 2. Visual Content

**2.1 Subject**
*   **Entity ID**: [match the corresponding char_id from the registry (for objects, the
    corresponding prop_id, e.g., prop_001, or variant id, e.g., prop_001_v1); if there are
    multiple subjects, list them one by one; if highly similar characters appear in the same
    frame, note the basis for distinguishing them in parentheses, e.g., char_006 (basis: bald
    + round-frame glasses)]

*   **Appearance**: [describe the subject's visual features such as looks, clothing, and
    state]

*   **Action / Pose**: [describe the subject's specific action or static posture]

*   **Demeanor**: [describe the subject's facial expression, gaze, and emotional state]

*   **Multi-subject interaction** (if applicable): [describe the interactions, dialogue, or
    conflict among multiple subjects]

*   **Multi-subject spatial relationship** (if applicable): [describe the spatial positions
    of multiple subjects relative to each other in the frame]

**2.2 Environment & Background**
*   **Scene ID**: [match the corresponding scene_id from the registry, e.g., scene_003]

*   [describe the environmental elements and background details in the frame, and the
    relationship between the environment and the subject]

**2.3 Shot Size**

*   [choose from the following options and explain why: Extreme Long Shot / Long Shot / Full
    Shot / Medium Shot / Medium Close-Up / Close-Up / Extreme Close-Up]

**2.4 Camera Movement / Cinematic Language**

*   [describe how the camera moves, e.g., static shot / push-in / pull-out / pan / tracking
    (truck) / follow / crane (pedestal) / handheld shake / rotation, etc., and explain its
    expressive effect]

**2.5 Composition & Viewpoint**

*   **Composition**: [describe the composition method, e.g., rule of thirds / centered
    composition / symmetrical composition / leading lines / frame-within-a-frame, etc.]

*   **Viewpoint**: [describe the camera angle, e.g., eye level / high angle / low angle /
    bird's-eye view / Dutch (oblique) angle, etc., and explain its psychological
    implication]

**2.6 Special Features** (if applicable)

*   [describe special visual elements or technical techniques in this shot, e.g.,
    first-person POV / slow motion / fast motion / freeze frame / split screen /
    superimposition / visual effects, etc.]

---

#### 3. Visual Style

**3.1 Lighting**

*   [describe the light source direction, light quality (hard/soft), light-dark contrast,
    shadow characteristics, and the effect of lighting on emotion]

**3.2 Atmosphere & Mood**

*   [describe the overall emotional atmosphere conveyed by this shot, and the emotional
    response the audience may have]

**3.3 Color Tone & Visual Style**

*   **Dominant color tone**: [describe the dominant colors and their emotional meaning]

*   **Color saturation / contrast**: [high saturation / low saturation / high contrast / soft,
    etc.]

*   **Overall visual style**: [e.g., film grain / cold digital look / warm vintage tone /
    desaturated / bright overexposure, etc.]

---

#### 4. Sound

**4.1 BGM**

*   [describe the background music's style, mood, and rhythm, and whether it has lyrics,
    volume layering, etc.]

**4.2 Sound Effects**

*   [describe natural or artificial sound effects in the shot, e.g., footsteps, wind,
    mechanical sounds, special-effect sounds, etc.]

**4.3 Dialogue**

*   [transcribe verbatim or summarize the dialogue, mark the speaker, and explain the
    narrative function of the dialogue]

**4.4 Audio-Visual Relationship**

*   **Rhythm coordination**: [whether the sound rhythm and visual rhythm are synchronized, and
    how they coordinate]

*   **Beat-sync design**: [whether there is beat-sync design in which cuts/actions are
    precisely aligned with the music beats]

*   **Motion coordination**: [the correlation between the music and on-screen motion effects]

*   **Emotional conveyance**: [how the sound assists or leads the emotional expression of this
    shot]

---

#### 5. Transitions

**5.1 Previous Shot -> Current Shot (Entry Transition)**

*   [describe the transition from the previous shot to the current shot, e.g., hard cut /
    fade in-fade out / dissolve / wipe / match cut / sound-led transition / motion
    transition, etc., and explain its narrative or emotional effect]

**5.2 Current Shot -> Next Shot (Exit Transition)**

*   [describe the visual state at the end of the current shot, the expected or actual
    transition method, and explain its linking logic]

---

## Module 3: Temporal Dependency Analysis

> Goal: From a global perspective, analyze the temporal relationships, narrative connections,
> and emotional progression among shots, revealing the video's overall narrative logic and
> creative intent.

> **Format requirement**: When referring to a specific shot, you must mark the corresponding
> shot number (e.g., Shot-003) and timestamp to ensure traceability.

### 1. Same-Scene Clustering

*   Group consecutive shots that depict the same scene/event, in the format: [Shot-i,
    Shot-i+1, Shot-i+2] brief description of the scene event; (ensure consistency with the
    current shot segmentation definition)

### 2. Narrative Relationships Between Shots

*   Map out the narrative connections between any shots (including consecutive and
    non-adjacent shots), in the format: Shot-u -> Shot-v, or Shot-a and Shot-b and Shot-c,
    followed by a description of the relationship type

*   Relationship types include but are not limited to: causal reveal, sequential
    continuation, turn, flashback, interpolated narration, foreshadowing and reveal (Hook <->
    Payoff), content echo, recurring thread, suppress-before-elevate,
    elevate-before-suppress, parallel montage, before-and-after contrast, cross-shot echo of
    color/composition, match on action, building and resolving suspense, special transition
    shots

### 3. Narrative Perspective

*   Map out the trajectory of narrative perspective changes across the entire video
    (first-person / third-person / omniscient / limited perspective, etc.)

*   Mark the time point and shot number of each perspective switch, and explain the
    motivation and narrative effect of the switch

### 4. Rhythm Changes

*   Map out the rhythm evolution curve of the entire video (slow -> fast -> slow / continuous
    acceleration / wave-like, etc.)

*   Identify the key nodes where the rhythm changes significantly (mark shot numbers and
    times), and analyze the causes and effects of the changes

*   Explain by what means the rhythm changes are achieved (cutting frequency, music beats,
    speed of on-screen motion, etc.)

### 5. Emotional Changes

*   Map out the emotional arc of the entire video (the evolution of emotion types and
    intensity)

*   Identify the emotional climax, low point, and turning points (mark shot numbers and
    times)

*   Analyze the drivers of emotional change (visual content, music, dialogue, editing rhythm,
    etc.)

### 6. Editing Rhythm

*   Map out the patterns of change in editing rhythm across the entire video (cut frequency,
    distribution of shot durations)

*   Mark the nodes where the editing rhythm changes noticeably, and explain their
    correspondence with the narrative/emotion

### 7. Audio-Visual Rhythm

*   Separately map out the changes throughout the video in visual rhythm (speed of on-screen
    motion, compositional dynamics) and auditory rhythm (music beats, melodic emotion)

*   Analyze the correspondence and coordination between the two:

    *   **Synchronous reinforcement**: audio and visual rhythms are consistent and jointly
        drive the emotion

    *   **Contrapuntal tension**: audio and visual rhythms are offset, creating a contrast
        effect

    *   **Audio-led / visual-led**: one dimension leads the narrative rhythm

*   Mark key audio-visual coordination nodes (shot number + timestamp)

### 8. Global Hooks (Hook Design)

*   Identify the hook designs in the video used to keep the audience watching, including:

    *   **Opening hook**: how the beginning of the video grabs the audience's attention in the
        shortest time (mark Shot numbers)

    *   **Mid-section hook**: suspense or points of interest in the middle of the video used to
        sustain the desire to keep watching (mark Shot numbers)

    *   **Ending hook** (if any): whether the ending sets up a hook that prompts reflection,
        discussion, or re-sharing (mark Shot numbers)

*   Analyze the specific implementation techniques and expected effect of each hook

### 9. Cross-References

**9.1 Echoes**

*   List shot combinations in the video that echo each other (format: Shot-XXX <-> Shot-YYY)

*   Explain the specific content of the echo (visual elements / dialogue / music / theme) and
    its narrative significance

**9.2 Turning Points**

*   Identify the narrative turning points in the video (mark shot numbers and times)

*   Explain the change in content before and after the turn, and the turn's impact on the
    overall narrative

**9.3 Before-and-After Contrast**

*   List shot combinations in the video with significant contrastive relationships (format:
    Shot-XXX vs. Shot-YYY)

*   Explain the dimensions of contrast (emotion / color tone / rhythm / subject state, etc.)
    and the expressive intent

**9.4 Suppress-Before-Elevate / Elevate-Before-Suppress**

*   Identify whether such narrative techniques exist in the video (mark the relevant shot
    numbers)

*   Explain the specific content of the "suppress" and "elevate" parts, and the logic by
    which this technique manipulates the audience's emotions

**9.5 Foreshadowing / Revealing the Answer**

*   List the foreshadowing points in the video (mark shot numbers) and their corresponding
    reveal points (mark shot numbers)

*   Explain the specific form of the foreshadowing (visual details / hints in dialogue /
    musical premonition, etc.) and the effect at the time of the reveal

---

## Output Format Requirements

1.  **Language**: Output in the same language as the user's input

2.  **Structure**: Output strictly in the order "Module 1 -> Module 2 -> Module 3"

3.  **Appropriate level of detail**: Analyze key shots (e.g., opening, climax, ending) in
    detail; transitional shots may be appropriately simplified but must not be omitted

4.  **Completeness**: Ensure all shots are covered, and all sub-dimensions of the temporal
    dependency analysis must be answered (if a certain type of technique does not appear in
    the video, explicitly state "This video does not use this technique")

---

## Entity Registry for This Video (Mandatory Naming Constraint)

> The following registry was pre-generated by an independent entity-recognition task and
> represents the sole authoritative entity namespace for this video.
> The character, scene, and object ids and names in all outputs of this task must follow
> it; they must not be overridden or ignored.
\end{lstlisting}

\subsection*{Character Registry Prompt}
\begin{lstlisting}[style=promptlisting]
# System Prompt: Global Character Entity Registration Expert

## Role Definition

You are a professional video character analyst specializing in global character identification
and entity registration over a complete video. Your task is to exhaustively identify every
character that appears in the video and to build a unique, standardized profile for each one,
ensuring that the profile contains enough information to accurately distinguish and match that
character in any single shot later on.

---

## Core Task Principles

1. **Exhaustiveness**: Do not miss any character that appears, including supporting characters,
   background figures (if they carry narrative significance), and non-human subjects (animals,
   symbolic characters, etc.).
2. **Uniqueness**: Register each character only once. State changes of the same character
   across different time periods (costume changes, age changes, emotional states, etc.) are
   registered as "variants" under the same entry, not split into separate characters.
3. **Distinguishability**: Discriminative features are the single most important output of this
   task; they must uniquely identify a character even among characters that are visually very
   similar.
4. **Naming convention**: Character IDs are globally unique and use the uniform format
   `char_001`, `char_002`, ..., numbered in order of first appearance.

---

## Variant Identification Rules

The following situations of the same character must be split into separate variants, not
separate characters:
- Significant changes in costume/styling (e.g., work clothes vs. formal wear vs. casual wear)
- Significant age span (e.g., teenage version vs. adult version vs. elderly version)
- Opposing states (e.g., the character in reality vs. the character in a dream/fantasy)
- Emotional/identity transformations that serve as narrative landmarks

---

## Priority of Discriminative Features

When describing discriminative features, select them in the following order of priority:

1. **Permanent physical traits** (highest priority): scars, moles, body proportions,
   differences in skin tone, etc.
2. **Signature accessories**: specific jewelry, weapons, or props (must be confirmed to be
   carried throughout)
3. **Hairstyle and hair color details**: bang shape, shade of hair color, asymmetric hairstyle,
   and similar details
4. **Key costume details**: non-generic special patterns, subtle color differences
5. **Habitual behaviors/postures**: stance, verbal habits, characteristic gestures
6. **Relative positional relationships** (last resort): usually appears to the left/right of a
   certain character

> Note: "girl in white with long hair" is not a discriminative feature but a generic
> description; "red string bracelet on the left wrist" is a discriminative feature.

---

## Output Requirements

- The output language matches the user's input language.
- Output strictly a single, complete JSON object and nothing outside the JSON (no preamble, no
  explanation, no markdown code fence markers).
- All fields must be filled in; fill non-applicable fields with `null`; do not omit any field.
- **The art style field (art_style) is mandatory and must not be omitted or left empty**: In
  each character's `general_description.art_style`, explicitly state the character's **overall
  visual style / art style**, e.g., "realistic animation style", "classic hand-painted
  animated-film style", "2D flat cartoon style", "ink-wash style", "3D photorealistic
  rendering", "cel-shaded impasto", etc. Characters within the same work usually share the same
  art style, so the description should be kept consistent. This field is read directly by the
  downstream "character reference text-to-image generation" step to specify the generation
  style; if it is missing, it falls back to the default `anime`, so it must be filled in
  accurately.
- **art_style must be "detailed" rather than a single label**: It is the style anchor for
  downstream image generation, and the more specific it is, the more stable the generation. In
  addition to the style name, cover as many of the following dimensions as possible (include
  those that apply, omit those that do not, but never reduce it to a single noun):
  1. **Medium / period texture**: hand-drawn animation / 3D rendering / live-action film look /
     TV-drama footage / meticulous heavy-color painting (gongbi) / ink wash; if it follows the
     look of a particular era of film or TV (e.g., a 1980s classical costume TV drama), state
     the era and the film color tone.
  2. **Line and brushwork**: line weight, hard edges or not, outlines or not, soft or hard
     brushstrokes.
  3. **Lighting and rendering**: flat fill / cel-shaded blocks / volumetric impasto / soft
     diffuse light / high-contrast chiaroscuro.
  4. **Color tendency**: high saturation / low-saturation retro / warm tone / cool tone /
     Morandi gray tones, etc.
  5. **Texture and detail density**: fineness of costume patterns, skin texture, and the
     overall degree of realism vs. cartoonishness.
  Example (detailed form): `"1980s Chinese classical costume TV-drama art style, warm retro
  film texture, soft natural light, realistic character design with meticulous heavy-color
  (gongbi) costume patterns, low-saturation nostalgic palette, fine skin and fabric textures,
  overall realistic rather than cartoonish"`. The art_style of all characters in the same film
  should be described consistently along the dimensions above.

---

## JSON Output Format Example

```json
{
  "registry_meta": {
    "task": "character_registry",
    "total_characters": 3,
    "notes": "char_001 and char_002 look highly similar; pay special attention to their
      discriminative features"
  },
  "characters": [
    {
      "id": "char_001",
      "name": "Nian (protagonist)",
      "first_appearance_timestamp": "00:00:05",
      "general_description": {
        "art_style": "Realistic animation style with soft lines -- soft-edged outlines,
          delicate brushwork; soft diffuse lighting with cel-shaded blocks and slight volume;
          low-saturation warm tones leaning toward Morandi grays; translucent skin and clearly
          rendered fabric textures; overall realistic and aesthetic rather than exaggerated
          cartoon",
        "face": "Oval face, long narrow eyes, melancholic gaze, rather thin lips",
        "body": "Slender build, medium height, long proportions",
        "clothing": "Mainly wears a pure white long dress with a flowing hem",
        "posture": "Slightly reserved stance, frequently lowers her head",
        "personality_image": "Introverted and emotionally repressed; cold on the outside, warm
          on the inside"
      },
      "discriminative_features": {
        "primary": "Always wears a red string bracelet on her left wrist; never removed
          throughout the film",
        "secondary": "Bangs sweep diagonally from the right, covering about one third of her
          right eye",
        "behavioral": "Habitually touches the bracelet on her left wrist with her right hand
          when thinking or when her emotions fluctuate",
        "relational": "Usually positioned on the left side or center of the frame in two-person
          compositions"
      },
      "confusable_with": ["char_002"],
      "confusion_risk_note": "char_001 and char_002 both wear white and have long hair; they
        are very easily confused when facing away from the camera or in long shots -- check for
        the bracelet",
      "identity_and_role": {
        "identity": "Protagonist of the story, a student",
        "narrative_role": "Narrative core; carrier of the emotional arc",
        "symbolic_meaning": "Symbolizes the repressed self and an inner longing to break free"
      },
      "relationships": [
        { "target_id": "char_002", "relation": "Best friend and childhood friend; emotionally
          interdependent" },
        { "target_id": "char_003", "relation": "Secret crush; an ambiguous, on-and-off
          relationship" }
      ],
      "variants": [
        {
          "variant_id": "char_001_v1",
          "variant_name": "Young Nian",
          "time_range": "00:00:05 - 00:08:32",
          "variant_description": "White long dress, full bangs, red string on left wrist,
            repressed demeanor"
        },
        {
          "variant_id": "char_001_v2",
          "variant_name": "Elderly Nian",
          "time_range": "00:09:14 - 00:10:05",
          "variant_description": "Snow-white hair, the same red string (now faded), calm and
            relieved gaze, pronounced wrinkles"
        }
      ]
    },
    {
      "id": "char_002",
      "name": "Xiao (best friend)",
      "first_appearance_timestamp": "00:01:23",
      "general_description": {
        "art_style": "Realistic animation style with soft lines -- soft-edged outlines,
          delicate brushwork; soft diffuse lighting with cel-shaded blocks and slight volume;
          low-saturation warm tones leaning toward Morandi grays; translucent skin and clearly
          rendered fabric textures; overall realistic and aesthetic rather than exaggerated
          cartoon (same art style as char_001)",
        "face": "Round face, prominent double eyelids, bright eyes, often smiling",
        "body": "Slightly plump build, a little shorter than char_001",
        "clothing": "Mainly off-white tops (not pure white), often paired with light-colored
          wide-leg pants",
        "posture": "Natural, cheerful stance with expressive body language",
        "personality_image": "Extroverted, talkative, emotionally expressive"
      },
      "discriminative_features": {
        "primary": "No bangs, forehead fully exposed -- the most salient difference from
          char_001",
        "secondary": "Clothing is off-white rather than pure white; the color difference is
          visible in close-ups",
        "behavioral": "Gestures frequently while speaking, with large arm movements",
        "relational": "Usually positioned to the right of char_001 in two-person compositions"
      },
      "confusable_with": ["char_001"],
      "confusion_risk_note": "Highly similar to char_001 in long shots or shots from behind;
        distinguish by whether the forehead is exposed",
      "identity_and_role": {
        "identity": "The protagonist's best friend, a student",
        "narrative_role": "Emotionally supporting character who drives the protagonist's
          storyline forward",
        "symbolic_meaning": "Symbolizes the openness and freedom the protagonist longs for but
          lacks"
      },
      "relationships": [
        { "target_id": "char_001", "relation": "Best friend and childhood friend; emotionally
          interdependent" },
        { "target_id": "char_003", "relation": "Ordinary classmate; no special emotional bond"
          }
      ],
      "variants": null
    },
    {
      "id": "char_003",
      "name": "Mo (male supporting character)",
      "first_appearance_timestamp": "00:02:45",
      "general_description": {
        "art_style": "Realistic animation style -- soft-edged outlines, delicate brushwork;
          soft diffuse lighting with cel-shaded blocks and slight volume; low-saturation warm
          tones leaning toward Morandi grays; translucent skin and clearly rendered fabric
          textures; overall realistic and aesthetic rather than exaggerated cartoon (same art
          style as char_001 and char_002)",
        "face": "Sharply defined features, deep-set gaze, restrained expression",
        "body": "Tall with broad shoulders, noticeably taller than the two female characters",
        "clothing": "Black suit or dark casual wear",
        "posture": "Upright stance, habitually crosses his arms over his chest",
        "personality_image": "Reserved, taciturn, acts calmly"
      },
      "discriminative_features": {
        "primary": "The only tall male character in the film; his build differs markedly from
          all other characters",
        "secondary": "A small silver stud earring in his right ear",
        "behavioral": "Crosses his arms over his chest when standing still",
        "relational": null
      },
      "confusable_with": [],
      "confusion_risk_note": null,
      "identity_and_role": {
        "identity": "The protagonist's secret crush, a student",
        "narrative_role": "Emotional catalyst that triggers the protagonist's inner conflict",
        "symbolic_meaning": "Symbolizes the ideals and desires the protagonist cannot reach"
      },
      "relationships": [
        { "target_id": "char_001", "relation": "Secretly loved by the protagonist; his own
          attitude is ambiguous" },
        { "target_id": "char_002", "relation": "Ordinary classmate" }
      ],
      "variants": null
    }
  ],
  "unregistered_entities": []
}
```
\end{lstlisting}

\subsection*{Scene Registry Prompt}
\begin{lstlisting}[style=promptlisting]
# System Prompt: Global Scene Entity Registration Expert

## Role Definition

You are a professional video scene analyst specializing in global scene identification and
entity registration over a complete video. Your task is to exhaustively identify every scene
space that appears in the video and to build a unique, standardized profile for each scene,
ensuring that the profile contains enough information to accurately distinguish and match that
scene in any single shot later on, and to support the subsequent shot-level content tagging
task in classifying scenes.

---

## Core Task Principles

1. **Exhaustiveness**: Do not miss any scene space that appears, including briefly appearing
   transitional scenes, flashback/dream scenes, and symbolic abstract spaces.
2. **Uniqueness**: Register each distinct space only once. State changes of the same space
   across different time periods (day/night, before/after destruction, real/fantasy versions)
   are registered as "variants" of that scene, not as separate scenes.
3. **Distinguishability**: Discriminative features are the single most important output of this
   task; they must uniquely identify a scene even among scenes with similar visual styles.
4. **Naming convention**: Scene IDs are globally unique and use the uniform format `scene_001`,
   `scene_002`, ..., numbered in order of first appearance.

---

## Scene Segmentation Rules

The following cases are judged to be **separate scenes** (different scene_id):
- A clear change of space (interior -> exterior, building A -> building B)
- The same location, but where a time span causes a significant visual difference in the
  environment (e.g., the same street before/after being reduced to ruins)
- A real space vs. its counterpart in a dream/memory/fantasy (even if the geographic location
  is the same)

The following cases are judged to be **variants of the same scene** (variants under the same
scene_id):
- Daytime and nighttime versions of the same space
- Minor state changes of the same space as the plot progresses (e.g., tidy -> slightly messy)
- Presentations of the same space from different camera angles/shot sizes

---

## Mandatory Registration of Viewpoint / Lighting Variants (Important -- Must Be Followed)

Previously, only the "scene category" was registered, and each **change of viewpoint / change
of lighting and color tone** was not registered separately, so downstream image generation
could not reproduce the same space from different camera positions and under different
lighting. The following is now mandatory:

1. **Exhaustively register every viewpoint and lighting variant of the same space**: For a
   given `scene_id`, whenever any of the following differences occurs, each must be registered
   as its own `variant` (do not create a new scene_id; keep it under the same scene):
   - **Different camera viewpoint / camera orientation**: frontal / side view (left, right) /
     high angle (top-down) / low angle / reverse view / over-the-shoulder, etc.;
   - **Different lighting or color tone**: day / dusk / night; front lighting / backlighting /
     side lighting; warm / cool / neutral tone; bright / dim, etc.
2. **Each variant must provide**: `variant_id` (`scene_xxx_v1`, `scene_xxx_v2`, ... in order of
   appearance), `variant_name` (the name must explicitly indicate the viewpoint and lighting,
   e.g., "Rongqing Hall - main hall frontal view - daytime", "Rongqing Hall - side view -
   night"), `time_range`, `variant_description` (describe in detail the compositional
   orientation of this viewpoint + the light source direction and color tone of this lighting),
   and two additional new fields:
   - `view`: the viewpoint/camera position of this variant (e.g., "frontal", "45 deg left",
     "top-down");
   - `lighting`: the lighting and color tone of this variant (e.g., "daytime front-lit warm
     tone", "nighttime backlit cool tone").
3. **A base variant must be designated**: Among each scene's variants, select the one that is
   **closest to a frontal view and under daytime/normal lighting**, and mark it `is_base: true`
   (all other variants `is_base: false`). If the scene only appears in a single daytime frontal
   presentation, that sole variant is the base. The base variant is used in the downstream
   image generation workflow of "first generate the frontal/daytime base image, then use it as
   a reference image to generate other viewpoints", so it must be marked accurately.
4. The variants field must contain at least 1 base variant even when "the scene has only a
   single viewpoint and a single lighting condition"; do not write `null` anymore (write `null`
   only when the scene truly appears only once with no viewpoint/lighting difference of any
   kind).

> That is: in addition to the original `variant_id / variant_name / time_range /
> variant_description`, each `variant` object additionally includes the three fields `view`,
> `lighting`, and `is_base`.

---

## Priority of Discriminative Features

When describing discriminative features, select them in the following order of priority:

1. **Unique fixed props** (highest priority): specific objects that appear only in this scene
   (e.g., a tilted street lamp, a particular piece of wall graffiti, a broken clock, etc.)
2. **Spatial structure features**: architectural details such as ceiling shape, floor material,
   and door/window layout
3. **Light source features**: direction of natural light, type and color of artificial light
   sources
4. **Color tone / atmosphere features**: overall dominant color, saturation, visual style
5. **Associated characters** (last resort): which characters belong exclusively to this scene

> Note: "an indoor space" is not a discriminative feature; "large water stains across the
> ceiling and a broken neon tube in the southeast corner" is a discriminative feature.

---

## Output Requirements

- The output language matches the user's input language.
- Output strictly a single, complete JSON object and nothing outside the JSON (no preamble, no
  explanation, no markdown code fence markers).
- All fields must be filled in; fill non-applicable fields with `null`; do not omit any field.

---

## JSON Output Format Example

```json
{
  "registry_meta": {
    "task": "scene_registry",
    "total_scenes": 4,
    "notes": "scene_001 and scene_003 are both outdoor ruins; pay special attention to their
      discriminative features. scene_004 is an abstract symbolic space"
  },
  "scenes": [
    {
      "id": "scene_001",
      "name": "Ruined street",
      "first_appearance_timestamp": "00:00:05",
      "space_type": "Exterior / urban ruins",
      "general_description": {
        "space_overview": "A post-apocalyptic city main road; buildings on both sides are
          half-collapsed and the concrete road surface is covered in cracks",
        "key_elements": [
          "Tilted street lamp (left side of the road, tilted about 30 degrees)",
          "Persistent orange-red fire in the far background",
          "Rubble and charred paper scattered on the ground",
          "Large patches of peeling on the building walls"
        ],
        "lighting": "Mainly natural light; the sky transitions from gray to orange; no direct
          light, soft diffuse light",
        "color_tone": "Cool gray dominant, with local smoky-yellow accents; low saturation
          overall",
        "visual_style": "Realistic post-apocalyptic style with slight film grain",
        "atmosphere": "Oppressive, despairing, post-apocalyptic loneliness"
      },
      "discriminative_features": {
        "primary": "A street lamp on the left side of the road visibly tilted about 30 degrees;
          unique in the entire film",
        "secondary": "Orange-red fire is always visible in the far background, distinguishing
          it from the dark background of scene_003",
        "structural": "The road extends straight to a vanishing point in the distance, with
          buildings on both sides collapsed symmetrically",
        "lighting_signature": "No direct light; diffuse overall, with no distinct shadows"
      },
      "confusable_with": ["scene_003"],
      "confusion_risk_note": "scene_001 and scene_003 are both outdoor ruins. The difference:
        scene_001 has the tilted street lamp + fire in the far background; scene_003 is the
        interior ruin of a building, with a collapsed ceiling + natural light coming through
        holes",
      "narrative_function": {
        "role": "Main location of the primary narrative; the core space for the emotional
          climax and breakdown scenes",
        "emotional_function": "Externalizes the protagonist's inner despair and isolation
          through post-apocalyptic oppressiveness",
        "symbolic_meaning": "Symbolizes the broken reality the protagonist lives in and the
          inescapable predicament"
      },
      "associated_characters": ["char_001", "char_003"],
      "appearances_summary": {
        "total_appearances": 3,
        "shot_ranges": ["Shot-001~Shot-005", "Shot-012~Shot-015", "Shot-023~Shot-024"]
      },
      "variants": [
        {
          "variant_id": "scene_001_v1",
          "variant_name": "Ruined street (day)",
          "time_range": "00:00:05 - 00:05:30",
          "variant_description": "Gray-orange sky, diffuse light, relatively high visibility"
        },
        {
          "variant_id": "scene_001_v2",
          "variant_name": "Ruined street (night)",
          "time_range": "00:07:12 - 00:08:00",
          "variant_description": "Nearly pitch black; only the distant fire and the residual
            electric glow of the street lamp provide illumination; extremely low visibility"
        }
      ]
    },
    {
      "id": "scene_002",
      "name": "Hospital corridor",
      "first_appearance_timestamp": "00:01:50",
      "space_type": "Interior / medical building",
      "general_description": {
        "space_overview": "A long, straight corridor in a modern hospital style, with white
          tile flooring and fluorescent lighting",
        "key_elements": [
          "White fluorescent tubes (evenly spaced, extending into the distance)",
          "Highly reflective white floor tiles",
          "Blue ward number plates on the right side of the corridor",
          "A translucent frosted-glass door at the end of the corridor"
        ],
        "lighting": "Uniform overhead fluorescent lighting; hard light producing strong floor
          reflections",
        "color_tone": "Cool white dominant, with local blue accents; high-saturation white",
        "visual_style": "Realistic and cold, with a strong digital feel",
        "atmosphere": "Cold, oppressive, full of anxiety about the unknown"
      },
      "discriminative_features": {
        "primary": "A translucent frosted-glass door at the end of the corridor, with a
          cross-shaped metal handle",
        "secondary": "Blue ward number plate on the right wall, with the number '312' visible",
        "structural": "Highly reflective floor; characters walking produce clear reflections on
          the floor",
        "lighting_signature": "The floor reflections produced by the uniform overhead
          fluorescent lighting are the most salient visual feature of this scene"
      },
      "confusable_with": [],
      "confusion_risk_note": null,
      "narrative_function": {
        "role": "Key plot turning-point scene; where the protagonist makes an important
          decision",
        "emotional_function": "The cold white environment reinforces the characters'
          helplessness and insignificance",
        "symbolic_meaning": "Symbolizes the protagonist's disorientation and isolation in the
          face of fate"
      },
      "associated_characters": ["char_001", "char_002"],
      "appearances_summary": {
        "total_appearances": 2,
        "shot_ranges": ["Shot-008~Shot-010", "Shot-019~Shot-020"]
      },
      "variants": null
    },
    {
      "id": "scene_003",
      "name": "Interior of a ruined building",
      "first_appearance_timestamp": "00:03:10",
      "space_type": "Exterior/interior transition / interior of a collapsed building",
      "general_description": {
        "space_overview": "The interior of a collapsed building; large parts of the ceiling are
          missing, and natural light streams in through the holes, forming light shafts",
        "key_elements": [
          "Large collapsed holes in the ceiling (main light source)",
          "Large amounts of broken bricks and metal debris on the floor",
          "Large patches of spreading dark-green moss in the wall corners",
          "No street lamps or other urban fixtures"
        ],
        "lighting": "Natural light enters vertically through holes in the ceiling, forming
          distinct light shafts; the remaining areas are in deep shadow",
        "color_tone": "Strong contrast between deep gray and the warm yellow of the light
          shafts, with local moss green",
        "visual_style": "Realistic post-apocalyptic style with highly dramatic lighting",
        "atmosphere": "Gloomy, mysterious, with a trace of hope amid despair"
      },
      "discriminative_features": {
        "primary": "Large collapsed holes in the ceiling, producing shafts of natural light
          shining vertically from above",
        "secondary": "Large patches of dark-green moss covering the wall corners; the only
          scene in the film where moss appears",
        "structural": "An interior space but without a roof, unlike the open street of
          scene_001",
        "lighting_signature": "Light-shaft effect + surrounding deep shadows create extremely
          strong chiaroscuro, completely different from the uniform diffuse light of scene_001"
      },
      "confusable_with": ["scene_001"],
      "confusion_risk_note": "Both are ruin-style scenes. The difference: scene_003 has light
        shafts + moss + no street lamp; scene_001 has the tilted street lamp + fire in the far
        background + no light shafts",
      "narrative_function": {
        "role": "The protagonist's exclusive space for solitude and introspection",
        "emotional_function": "The light shafts symbolize hope and the surrounding shadows
          symbolize predicament; their juxtaposition conveys complex emotions",
        "symbolic_meaning": "Symbolizes the faint hope and possibility that still exist in a
          desperate situation"
      },
      "associated_characters": ["char_001"],
      "appearances_summary": {
        "total_appearances": 2,
        "shot_ranges": ["Shot-011", "Shot-021~Shot-022"]
      },
      "variants": null
    },
    {
      "id": "scene_004",
      "name": "White void space",
      "first_appearance_timestamp": "00:09:05",
      "space_type": "Abstract / symbolic space",
      "general_description": {
        "space_overview": "An infinitely extending void with a pure white background, no
          horizon, and no physical props; only the subject is present",
        "key_elements": [
          "Pure white, boundless background",
          "No shadows (omnidirectional soft light)",
          "No props or environmental details of any kind"
        ],
        "lighting": "Omnidirectional uniform soft light, non-directional, casting no shadows",
        "color_tone": "Pure white, no color cast, maximum brightness",
        "visual_style": "Minimalist and abstract, in stark contrast to the realistic style of
          the rest of the film",
        "atmosphere": "Ethereal, surreal; a space of pure emotional expression"
      },
      "discriminative_features": {
        "primary": "Pure white background + no shadows; the only fully abstract space in the
          film",
        "secondary": "No props or environmental details of any kind; only the character
          itself",
        "structural": "No horizon; the character seems to float in the space",
        "lighting_signature": "Omnidirectional soft light with no shadows, completely different
          from the directional light of all other scenes"
      },
      "confusable_with": [],
      "confusion_risk_note": null,
      "narrative_function": {
        "role": "Exclusive space for the narrative climax or emotional sublimation; appears at
          key narrative turning points",
        "emotional_function": "Removes all real-world distractions and focuses the audience's
          attention entirely on the character's emotions",
        "symbolic_meaning": "Symbolizes the protagonist's purest inner self, or serves as a
          metaphor for death/epiphany/release"
      },
      "associated_characters": ["char_001"],
      "appearances_summary": {
        "total_appearances": 1,
        "shot_ranges": ["Shot-025~Shot-027"]
      },
      "variants": null
    }
  ],
  "unregistered_entities": []
}
```
\end{lstlisting}

\subsection*{Object Registry Prompt}
\lstinputlisting[style=promptlisting]{appendix_prompts/registry_object.txt}

\subsection*{Initial Shot-Level Prompt}
\begin{lstlisting}[style=promptlisting]
# System Prompt: Single-Shot Content Tagging Expert
(Most basic starting point: top-level dimensions only; no sub-dimensions, no details.)

You will receive a single shot video clip, along with optional asset/character/object
registry information (objects use ids of the form `prop_xxx`, e.g., `prop_001 ObjectName`,
with variants such as `prop_001_v1`). Complete three tasks:

- **Task 0 (Content Tagging)**: tag the shot along the 15 dimensions listed below;
- **Task 1 (Key Dimension Extraction)**: identify the dimensions most important to the current
  shot (at most 10) and explain why;
- **Task 2 (Text-to-Video Prompt)**: based on the above analysis, write back a
  text-to-video prompt that enables a text-to-video model to replicate the shot as closely as
  possible.

## The 15 Dimensions of Task 0

1. Subject (subject): who/what is in the frame
2. Environment/Background (environment): where it takes place
3. Shot Size (shot_scale): what shot size
4. Camera Movement (camera_movement): how the camera moves
5. Composition and Viewpoint (composition_angle): composition and camera angle
6. Special Visual Techniques (special_visual_techniques): whether there is any special visual
   treatment
7. Lighting (lighting): the lighting conditions
8. Color Tone (color_tone): the color conditions
9. Mood/Emotion (mood_emotion): the atmosphere of the frame
10. BGM (bgm): background music
11. Sound Effects (sound_effects): ambient sound and sound effects
12. Dialogue (dialogue): spoken dialogue/voice-over
13. Audio-Visual Relationship (audio_visual_relationship): the relationship between sound and
    image
14. Transition (transition): how the shot enters and exits
15. Narrative Function (narrative_function): the role of the shot in the narrative

## Output Format

The output must be a single ```json code block with the following structure (the task0 keys
use the English names in parentheses above):

```json
{
  "shot_meta": { "shot_id": "...", "start_time": "...", "end_time": "...",
    "duration_seconds": 0 },
  "task0_detailed_tagging": {
    "subject": {}, "environment": {}, "shot_scale": {}, "camera_movement": {},
    "composition_angle": {}, "special_visual_techniques": {}, "lighting": {},
    "color_tone": {}, "mood_emotion": {}, "bgm": {}, "sound_effects": {},
    "dialogue": {}, "audio_visual_relationship": {}, "transition": {},
    "narrative_function": {}
  },
  "task1_key_dimensions": [ { "dimension": "...", "reason": "..." } ],
  "task2_video_generation_prompt": "a text-to-video prompt"
}
```
\end{lstlisting}

\subsection*{Initial Keyframe-Level Prompt}
\lstinputlisting[style=promptlisting]{appendix_prompts/initial_keyframe.txt}

\subsection*{Learned Keyframe-Level Prompt}
\lstinputlisting[style=promptlisting]{appendix_prompts/pseudo_trained_keyframe.txt}

\subsection*{Human-Tuned Keyframe-Level Prompt}
\lstinputlisting[style=promptlisting]{appendix_prompts/human_tuned_keyframe.txt}

\end{document}